%% file: main.tex
\documentclass[10pt,twocolumn,letterpaper]{article}

\usepackage{cvpr}              %

\input{preamble}

\def\paperID{6518} %
\def\confName{CVPR}
\def\confYear{2025}

\newcommand{\notsosmall}{\fontsize{10.5pt}{12pt}\selectfont}

\definecolor{somegray}{rgb}{0.5, 0.5, 0.5}
\newcommand{\darkgrayed}[1]{\textcolor{somegray}{#1}}
\makeatletter
\newcommand*\titleheader[1]{\gdef\@titleheader{#1}}
\AtBeginDocument{%
  \let\st@red@title\@title
  \def\@title{%
    \vskip-3em
    \bgroup\normalfont\large\centering\@titleheader\par\egroup
    \vskip1.5em\st@red@title}
}
\makeatother

\titleheader{\darkgrayed{This paper has been accepted for publication at the \\
IEEE Conference on Computer Vision and Pattern Recognition (CVPR), Nashville, 2025.
\copyright IEEE}\vspace{-0.3cm}}

\title{\includegraphics[scale=0.05]{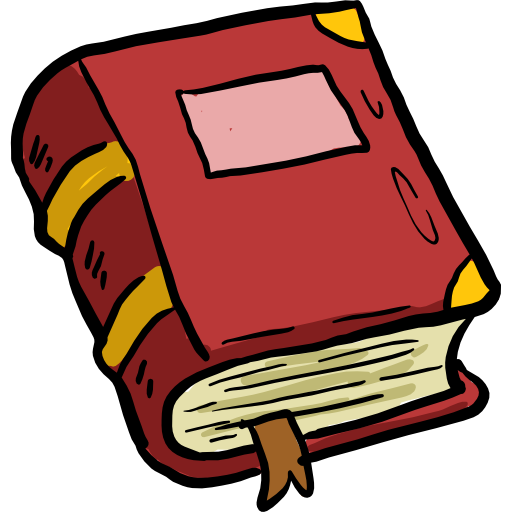} Semantic Library Adaptation: \\ LoRA Retrieval and Fusion for Open-Vocabulary Semantic Segmentation}

\author{Reza Qorbani$^{3}$\footnote{} \hspace{1cm} 
Gianluca Villani$^{1,2}$$^{*}$ \hspace{1cm} 
Theodoros Panagiotakopoulos$^{1,5}$ \\
Marc Botet Colomer$^{1}$ \hspace{1cm}
Linus Härenstam-Nielsen$^{6,7}$ \hspace{1cm}
Mattia Segu$^{1,9}$ \hspace{1cm} 
Pier Luigi Dovesi$^{1,4}$\footnote{} \\
Jussi Karlgren$^{1,4}$ \hspace{1cm} 
Daniel Cremers$^{6,7}$ \hspace{1cm} 
Federico Tombari$^{6,8}$ \hspace{1cm} 
Matteo Poggi$^{10}$ \vspace{0.3cm}\\
\notsosmall $^1$The Good AI Lab \hspace{1cm} 
$^2$University of Toronto \hspace{1cm}
$^3$KTH \hspace{1cm}
$^4$AMD Silo AI\\
\notsosmall $^5$King \hspace{1cm}
$^6$Technical University of Munich \hspace{1cm} 
$^7$Munich Center for Machine Learning \\
\notsosmall $^8$Google \hspace{1cm}
$^9$ETH Zurich \hspace{1cm}
$^{10}$University of Bologna \\
\small{\projectpage}\vspace{-0.5cm}
}

\begin{document}

\maketitle
\input{sec/0_abstract}    
\input{sec/1_intro}

\input{sec/2_relatedwork}
\input{sec/3_method}

\input{sec/4_experiments}
\input{sec/5_conclusions}

{
    \small
    \bibliographystyle{ieeenat_fullname}
    \bibliography{main,morebibs,egbib}
}

\input{sec/X_suppl}
\end{document}

%% file: preamble.tex
\definecolor{cvprblue}{rgb}{0.21,0.49,0.74}
\usepackage[pagebackref,breaklinks,colorlinks,allcolors=cvprblue]{hyperref}
\usepackage{overpic}
\usepackage{xcolor}
\usepackage{float}
\definecolor{forestgreen}{rgb}{0.13, 0.55, 0.13}
\definecolor{burgundy}{rgb}{0.5, 0.0, 0.13}
\usepackage{colortbl,xcolor}

\newcommand{\rem}[3]{%
    {\colorbox{#2}{\sffamily\scriptsize\bfseries\textcolor{white}{#1}}}
    {\sffamily\small\itshape\textcolor{#2}{#3}}
}

\usepackage{xcolor}

\definecolor{piergreen}{RGB}{30, 120, 30}
\definecolor{lightgray}{RGB}{200, 200, 200}

\newcommand{\methodname}[0]{SemLA\xspace}

\newcommand{\projectpage}[0]{\url{https://thegoodailab.org/semla}}
\usepackage{multirow}

\newcommand\blfootnote[1]{%
  \begingroup
  \renewcommand\thefootnote{}\footnote{#1}%
  \addtocounter{footnote}{-1}%
  \endgroup
}

%% file: sec/0_abstract.tex
\begin{abstract}

Open-vocabulary semantic segmentation models associate vision and text to label pixels from an undefined set of classes using textual queries, providing versatile performance on novel datasets. However, large shifts between training and test domains degrade their performance, requiring fine-tuning for effective real-world applications. We introduce \textbf{Semantic Library Adaptation (\methodname)}, a novel framework for training-free, test-time domain adaptation. \methodname leverages a library of LoRA-based adapters indexed with CLIP embeddings, dynamically merging the most relevant adapters based on proximity to the target domain in the embedding space. This approach constructs an ad-hoc model tailored to each specific input without additional training. Our method scales efficiently, enhances explainability by tracking adapter contributions, and inherently protects data privacy, making it ideal for sensitive applications. Comprehensive experiments on a 20-domain benchmark built over 10 standard datasets demonstrate \methodname’s superior adaptability and performance across diverse settings, establishing a new standard in domain adaptation for open-vocabulary semantic segmentation.

\end{abstract}

%% file: sec/1_intro.tex
\section{Introduction}
\label{sec:intro}

\blfootnote{$^*$ Joint first authorship \hspace{1cm} $^\dagger$ Project Lead}Semantic segmentation aims to classify images at the pixel level, assigning a label to each pixel in an image. Traditionally, models are trained to recognize a fixed set of classes, but recent advances have led to \emph{open-vocabulary} (OV) semantic segmentation, where models identify categories from an undefined and unbounded set using textual queries. This task leverages the synergy between text and visual embeddings~\cite{radford2021learning}, enabling flexible and dynamic labeling~\cite{Cho_2024_CVPR,xie2024sed}.
However, like all models, OV semantic segmentation models are vulnerable to \emph{domain shifts} -- situations where the model encounters data distributions different from those it was trained on, leading to degraded performance. This issue is significant because it impacts both segmentation accuracy and open-vocabulary generalization capabilities, which are crucial in real-world applications.

\begin{figure}
    \centering
    \includegraphics[trim=0cm 0cm 0cm 0cm, clip, width=0.4\textwidth]{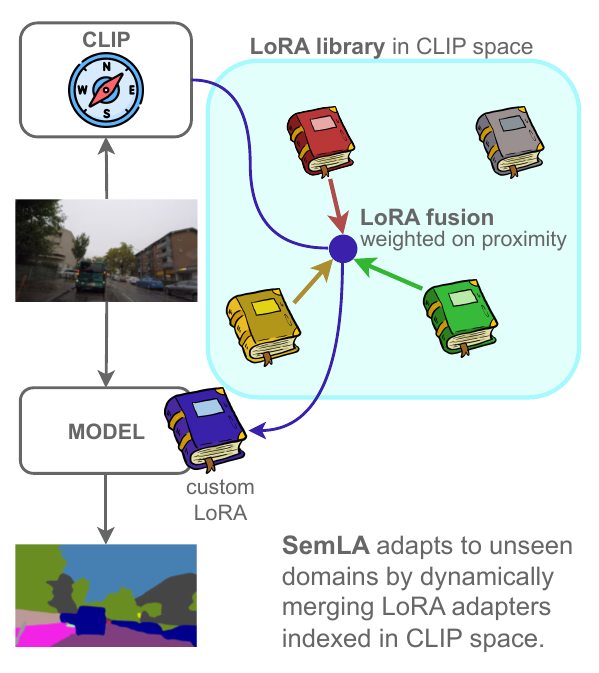}
    \vspace{-0.2cm}
    \caption{\textbf{Overview of \methodname}. During test-time, \methodname uses CLIP as a domain navigator, to retrieve and fuse relevant adapters, to get a LoRA tailored to the target domain.}
    \label{fig:cover}\vspace{-0.3cm}
\end{figure}

Data is plentiful but heterogeneous -- varying in labels, annotation styles, and other factors -- and often contains sensitive information, limiting its accessibility. Consequently, we find ourselves in a paradox where an abundance of data does not necessarily translate into robust models capable of handling domain shifts~\cite{wortsman2022robust}. Although several domain adaptation methods have been proposed, they have not been applied to OV semantic segmentation settings.

Classic domain adaptation approaches are limited: they usually focus on a single target domain, require access to source data during adaptation, involve slow and expensive processes, and often deteriorate performance on the source dataset~\cite{cyclegan,dcan,cycada,ganin,fada}. Test-time and online adaptation address some issues but are usually too slow for real-time applications and lack explainability~\cite{Panagiotakopoulos_ECCV_2022,colomer2023toadapt}.
Drawing inspiration from recent developments in Large Language Models (LLMs) and the use of \emph{Parameter-Efficient Fine-Tuning} (PEFT) techniques enriched with metadata -- such as adapters in Hugging Face and Civit-ai -- we aim to transfer this idea to the field of OV semantic segmentation, %
by integrating external information beyond what was found in the static training data through retrieval mechanisms \cite{loraretriever}. %

We propose a training-free test-time adaptation approach for OV semantic segmentation, \textbf{\textit{Sem}}antic \textbf{\textit{L}}ibrary \textbf{\textit{A}}daptation (\textbf{\methodname}), achieved through adapter retrieval and merging. Our approach begins with creating a set of adapters using Low-Rank Adaptation (LoRA)~\cite{hu2022lora,peft}.
However, simply having a collection of adapters is insufficient, as the model still can't discern which adapter to select. Just as a collection of books requires an indexing system to become a functional library, we introduce an indexing mechanism to guide retrieval. Specifically, we employ Contrastive Language-Image Pretraining (CLIP)~\cite{radford2021learning} to represent each LoRA adapter as the centroid of the CLIP embeddings of its training data. At test time, we can then select the most relevant adapters based on the proximity of the target image embedding to these centroids.

    Real-world domains are, however, far too numerous to rely on single adapters for every possible scenario due to the inherent combinatorial complexity of the domain space. To address this, 
    we aggregate the most relevant adapters based on their proximity to the target domain. Just as in a library where the exact book you seek may not be available, the necessary knowledge can still be gathered by consulting multiple related sources. By fusing several adapters based on their relevance to the target image, we effectively construct an ad-hoc model fine-tuned for each situation.  Figure \ref{fig:cover} presents an overview of \methodname at test-time.

This is training-free, scales well to any library size, and accommodates vastly different domains and label sets, with adaptation happening almost instantaneously through CLIP inference and LoRA merging.
Moreover, our approach enhances explainability by identifying which adapters are most useful for certain target domains, allowing us to understand the model's reasoning and capabilities. %

Our main contributions are summarized as follows:
\begin{itemize}
    \item \textbf{Training-free test-time adaptation for OV semantic segmentation}: We introduce the first method for adaptation in OV segmentation. It requires no training at test time, enabling adaptive responses to diverse input images. Our framework is simple, scalable with large adapter libraries, and backbone-agnostic.
    \item \textbf{Novel benchmark for OV domain adaptation}: We provide comprehensive experiments with a 20-domain benchmark, built over 10 popular datasets, showcasing performances over vastly different datasets and superior performance compared to zero-shot and naive adapter merging \cite{li2024training} approaches.
    \item \textbf{Explainability and LoRA contribution analysis}: Our approach is inherently transparent and controllable even at test time, easily scaling to new targets. The adaptation phase occurs without accessing source data, thereby protecting data privacy. We present analyses showing how adapters synergize and how we can measure their contributions and influence over the adaptation process.
\end{itemize}

%% file: sec/2_relatedwork.tex
\section{Related Work}

Our work intersects the following research areas:

\textbf{Semantic Segmentation.} Deep learning techniques have led to increasingly effective semantic segmentation models. Fully Convolutional Networks (FCN)~\cite{fcn} and SegNet~\cite{badrinarayanan2017segnet} extended convolutional neural networks with upsampling layers (deconvolution) to produce pixel-wise predictions. Subsequent works improved both speed~\cite{yu2018bisenet,nekrasovlight} and accuracy~\cite{chen2017deeplab,deeplabv2,chen2018encoder}. Enhancements in accuracy involved enlarging the receptive field~\cite{zhao2017pspnet,yang2018denseaspp,chen2017deeplab,deeplabv2,chen2018encoder}, designing refinement modules~\cite{fu2019adaptive,zhou2019context,he2019adaptive}, introducing boundary cues~\cite{chen2016semantic,ding2019boundary,takikawa2019gated}, and exploiting attention mechanisms~\cite{fu2019dual,li2019expectation,wang2018non} and Vision Transformers~\cite{xie2021segmenting,yuan2021segmentation,xie2021segformer}.

With the introduction of vision-language models like CLIP~\cite{radford2021learning}, \emph{open-vocabulary} (OV) semantic segmentation~\cite{Cho_2024_CVPR,xie2024sed} has emerged as a new approach to segmentation, where the set of classes can be arbitrarily defined at any time through textual queries. This flexibility, however, introduces additional sources of domain shifts, such as vocabulary misalignment across different domains.

\textbf{Domain Adaptation.} Domain adaptation focuses on adapting a model pre-trained on a source domain to perform well on a new, unlabeled target domain~\cite{zhang2021survey}. Adaptation can be performed either offline or during deployment. In the offline case, usually known as \emph{unsupervised domain adaptation} (UDA), early approaches used style transfer strategies~\cite{cyclegan,dcan,cycada,bdl,stylization,yang_fda_2020} or self-training to adapt the model, for example by carefully retrieving reliable pseudo-labels for the target domain by exploiting the confidence of the model itself~\cite{cbst,iast,zou2019confidence}, class balancing~\cite{zou2018unsupervised,hoyer2021daformer}, or prototypes~\cite{chen2019progressive,zhang_category_2019,zhang_prototypical_2021}. While the offline scenario usually focuses on domain shifts occurring only once, such as moving from synthetic~\cite{gta} to real~\cite{Cityscapes} images, the deployment approach aims to handle \emph{continuous} adaptation to avoid catastrophic forgetting of the source domain.
Some methods assume availability of data from the training domain and involve replay buffers~\cite{bobu_adapting_2018,lao2020continuous,kuznietsov2022towards}, style transfer~\cite{wu_ace_2019,fan2022normalization,fan2023towards}, contrastive learning~\cite{su_gradient_2020,vs2023towards}, or adversarial learning~\cite{wulfmeier_incremental_2018}. Others consider a more constrained case, \emph{source-free} or \emph{test-time} adaptation, where no data from the source domain is available~\cite{Stan2021UnsupervisedMA}, and involve pseudo-source data generation~\cite{liu_source-free_2021}, partial freezing of the original model~\cite{liang_we_2020}, feature alignment~\cite{liu2021ttt}, batch norm retraining through entropy minimization~\cite{wang2021tent}, or prototype adaptation~\cite{iwasawa2021testtime}. Recently, \emph{online} adaptation emerged to tackle multiple domain shifts occurring unpredictably during deployment~\cite{shift,Panagiotakopoulos_ECCV_2022,Volpi_2022_CVPR,colomer2023toadapt,segu2023darth}. Although these strategies are flexible for unseen domains, they introduce significant overhead at deployment.
Finally, \emph{training-free} adaptation~\cite{segu2023batch,li2024training} has been proposed to address domain shifts by aggregating parameters of different single-domain experts. This approach relies on linear mode connectivity to uniformly merge different models. We demonstrate that a careful selection of a few meaningful experts allows for better results.

\textbf{Parameter-Efficient Fine-Tuning.} The advent of large language models (LLMs) and vision-language models (VLMs), with billions of parameters trained on vast datasets, unveiled the need for new fine-tuning paradigms to adapt them to specific use cases. Low-Rank Adaptation (LoRA)~\cite{hu2021lora} was proposed to fine-tune LLMs by optimizing new sets of low-rank weights to learn residuals over the original parameters, preserving the knowledge derived from large-scale datasets while reducing computational load during fine-tuning. Advanced strategies allow dynamically setting the rank of LoRA parameters~\cite{zhang2023adaptive}. Recently, the possibility of merging different LoRAs learned for different tasks has gained popularity~\cite{huang2023lorahub}, yielding various policies for combining multiple LoRAs~\cite{peft}.

\textbf{Model Merging.} Model merging has emerged as a strategy to fuse knowledge from multiple sources directly in the weight space, effectively representing an alternative to ensemble and federated learning~\cite{deepmodelfusion}. In~\cite{wortsman2022robust}, merging parameters of zero-shot and fine-tuned models improves out-of-distribution performance. Similarly, \cite{ilharco2022patching,modelstock} use linear interpolation for multi-task learning. While these methods interpolate all weights of the models, our method merges LoRA adapters, which are orders of magnitude smaller in size. Inspired by Model Soups~\cite{modelsoups}, AdapterSoup~\cite{adaptersoup} averages adapters in parameter space, automatically retrieving and uniformly merging them for adaptation to downstream language tasks. LoraRetriever~\cite{loraretriever} introduced a similar pipeline but instead uses instruction fine-tuning to train a retriever for selecting the most relevant adapters. Compared to these methods, \methodname performs adapter retrieval and fusion simply relying on the powerful semantic indexing capabilities of CLIP~\cite{radford2021learning}, which effectively serves as a \emph{domain navigator}.

\textbf{Positioning Our Work.} Our framework falls into the category of \emph{domain adaptation} methods, as it actively adapts to new, unseen domains by modifying the model weights based on the input. However, we diverge from classical UDA, as our source dataset isn't fixed but can be changed at test time. It is represented by the data used to train the model backbone and all the datasets employed in the training of the adapters composing our LoRA library. These effectively represent the points in the LoRA weight-space over which we interpolate to generate new models for every unseen target image.

%% file: sec/3_method.tex
\section{Method and Framework}

In this section, we introduce \methodname. It consists of two main stages: (1) \textit{Construction of the LoRA Adapters Library}, and (2) \textit{Dynamic Test-Time Adaptation}. We begin by detailing the backbone architecture and the integration of Low-Rank Adaptation (LoRA) into it, followed by the description of our library creation and the adaptation process.

\begin{figure*}[t]
    \centering
    \includegraphics[trim=0.5cm 0cm 0.5cm 0cm, clip,width=0.8\textwidth]{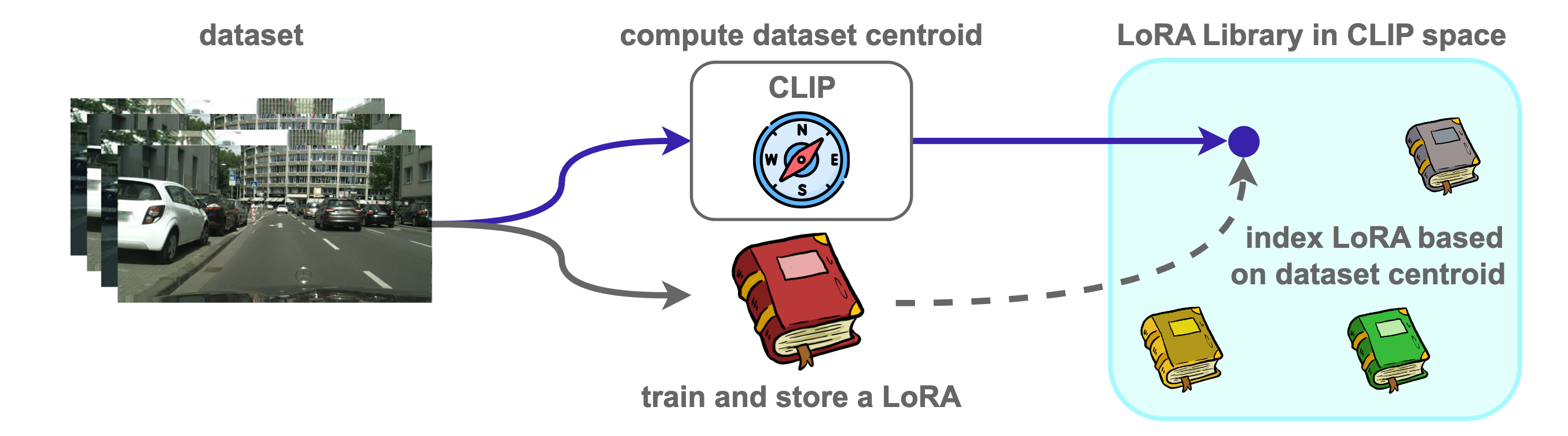}\vspace{-0.3cm}
    \caption{\textbf{Construction and Expansion of the LoRA Adapter Library.} Each LoRA adapter is created by fine-tuning on a specific dataset and subsequently added to the library. The library index for each adapter is represented by the CLIP centroid of its training data.}
    \label{fig:library_construction}\vspace{-0.3cm}
\end{figure*}

\subsection{Backbone Architecture with LoRA Integration}

Our framework builds on the \textbf{CAT-Seg} architecture~\cite{Cho_2024_CVPR}, a state-of-the-art model designed for OV semantic segmentation. It aligns visual and textual information by mapping images and textual queries into a shared semantic space using CLIP embeddings~\cite{radford2021learning}. This allows the model to perform segmentation without being constrained to a predefined set of classes, enabling flexible and dynamic labeling. \methodname can be deployed with any model enjoying these properties -- thus not only CAT-Seg, as we will discuss later.
 
To facilitate efficient and adaptable domain-specific tuning, we integrate \textbf{Low-Rank Adaptation (LoRA)}~\cite{hu2022lora} into CAT-Seg. LoRA introduces learnable low-rank matrices to each linear layer, allowing for parameter-efficient fine-tuning without altering the original weights. Specifically, for each linear layer with weights $\mathbf{W} \in \mathbb{R}^{d \times k}$, we augment it with low-rank matrices $\mathbf{B} \in \mathbb{R}^{d \times r}$ and $\mathbf{A} \in \mathbb{R}^{r \times k}$, where $r \ll \min(d, k)$. The adapted weights become:
\begin{equation}
\mathbf{W}' = \mathbf{W} + \Delta\mathbf{W}, \quad \Delta\mathbf{W} = \mathbf{B}\mathbf{A}.
\end{equation}
By training only the LoRA parameters $(\mathbf{B}, \mathbf{A})$ for each domain, we create lightweight domain-specific adapters without the need to retrain the entire model. For consistency and to simplify the merging process, we use the same rank $r$ for all LoRA adapters.

Moreover, we denote a LoRA adapter with $\Delta \mathcal{W}_i = (\mathcal{B}_i, \mathcal{A}_i)$, where $\mathcal{B}_i = \{B_{1, i}, \dots, B_{m, i}\}$ and $\mathcal{A}_i = \{A_{1, i}, \dots, A_{m, i}\}$ denote all the LoRA parameters applied across all $m$ linear layers of the CAT-Seg model. Given a LoRA adapter $\Delta \mathcal{W}_i$, the corresponding adapted CAT-Seg model is denoted by $\text{CAT-Seg}(\Delta \mathcal{W}_i)$, or $\text{CAT-Seg}(\Delta \mathcal{W}_i, \mathbf{x}_t)$ if we wish to specify the dependence on the input image $\mathbf{x}_t$. In the following, we explain the procedure for a single linear layer; updates for all other linear layers are performed analogously.

\subsection{Construction of the LoRA Adapters Library}

The first stage involves building a library of domain-specific LoRA adapters, each associated with a representative embedding in the CLIP space. This library serves as the foundation for our dynamic adaptation at test time. This stage, depicted in Figure \ref{fig:library_construction}, is performed offline and can be repeated for each newly labeled domain that becomes available for training.

\subsubsection{Domain Embeddings from CLIP}

For each training dataset (domain) $\mathcal{D}_i$, we compute a centroid embedding $\mathbf{c}_i$ to represent the domain in the CLIP embedding space as follows:

\begin{enumerate}[label=(\alph*)]
    \item \textbf{Embedding Computation}: For each image $\mathbf{x}_j \in \mathcal{D}_i$, we obtain its CLIP embedding $\mathbf{e}_j$:
    \begin{equation}
    \mathbf{e}_j = \text{CLIP}_{\text{image}}(\mathbf{x}_j).
    \end{equation}
    \item \textbf{Centroid Calculation}: We calculate the centroid of the embeddings from the $N_i$ images in $\mathcal{D}_i$:
    \begin{equation}
    \mathbf{c}_i = \frac{1}{N_i} \sum_{j=1}^{N_i} \mathbf{e}_j,
    \end{equation}
\end{enumerate}
The centroid $\mathbf{c}_i$ captures the semantic essence of $\mathcal{D}_i$, facilitating efficient similarity comparisons during adaptation.

\subsubsection{Training Domain-Specific LoRA Adapters}

For each domain $\mathcal{D}_i$, we fine-tune the LoRA parameters $\Delta \mathcal{W}_i = (\mathcal{B}_i, \mathcal{A}_i)$ on CAT-Seg using $\mathcal{D}_i$, while keeping the original model weights $\mathbf{W}$ frozen. This results in a domain-specific adapter $\text{LoRA}_i = \Delta \mathcal{W}_i$, which, when combined with the centroid $\mathbf{c}_i$, forms a tuple $(\mathbf{c}_i, \Delta \mathcal{W}_i)$ representing domain $\mathcal{D}_i$ in our library.

\subsubsection{Assembly of the Adapters Library}

By collecting all domain-specific adapters and their centroids, we construct the \textbf{LoRA Adapters Library}:

\begin{equation}
\mathcal{L} = \left\{ \left( \mathbf{c}_i, \Delta \mathcal{W}_i  \right) \mid i = 1, 2, \dots, M \right\},
\end{equation}
where $M$ is the total number of training domains. 
This library enables the model to dynamically adapt to various domains by selecting and merging relevant adapters.  

\subsubsection{Extending the Library}

Our approach is extremely scalable. In any point in the future, as soon as new annotated data become available for a new domain $\mathcal{D}_*$, we can perform steps 3.2.1 and 3.2.2 to obtain $(\mathbf{c}_*, \Delta \mathcal{W}_*)$ and append it to $\mathcal{L}$ to enrich it 
\begin{equation}
\mathcal{L} = \mathcal{L} + (\mathbf{c}_*, \Delta \mathcal{W}_*)
\end{equation}

\subsection{Dynamic Test-Time Adaptation}

The second stage focuses on adapting the model for each input image at test time by selecting and merging the most relevant adapters from the library, guided by CLIP.

\subsubsection{Computing the Test Image Embedding}

Given a test image $\mathbf{x}_t$, we compute its CLIP embedding $\mathbf{e}_t$:

\begin{equation}
\mathbf{e}_t = \text{CLIP}_{\text{image}}(\mathbf{x}_t).
\end{equation}
This embedding serves as a query to identify the most relevant adapters in the library.

\subsubsection{Selecting Relevant Adapters}

We measure the similarity between the test image embedding and each domain centroid, e.g. as Euclidean distance:

\begin{equation}
\label{eqn:adapter-selection}
d_i = \left\| \mathbf{e}_t - \mathbf{c}_i \right\|_2, \quad \forall (\Delta \mathcal{W}_i, \mathbf{c}_i) \in \mathcal{L}.
\end{equation}
We then select the top-$K$ adapters with the smallest distances, denoted by the index set $\mathcal{K} = \{ i_1, i_2, \dots, i_K \}$.

\subsubsection{Computing Adapter Weights}

To quantify the relevance of each selected adapter, we apply a softmax function with temperature $\tau$ to the proximities, computed as distance reciprocals:

\begin{equation}
w_i = \frac{\exp\left( \dfrac{1}{d_i \cdot \tau} \right)}{\sum_{k \in \mathcal{K}} \exp\left( \dfrac{1}{d_k \cdot \tau} \right)}, \quad \forall i \in \mathcal{K}.
\end{equation}
These weights $w_i$ determine the contribution of each adapter in the fusion process.

\subsubsection{Merging Adapters via Concatenation}

We specify here the merging procedure for a given linear layer adapter \cite{peft}. Since all adapters share the same rank $r$, the merging is straightforward:

    \begin{equation}
    \mathbf{A}_i' = w_i \mathbf{A}_i, \quad \forall i \in \mathcal{K}.
    \end{equation}
    \begin{equation}
    \mathbf{A}_{\text{fused}} = \left[ \mathbf{A}_{i_1}'^\top, \mathbf{A}_{i_2}'^\top, \dots, \mathbf{A}_{i_K}'^\top \right]^\top \in \mathbb{R}^{rK \times d}.
    \end{equation}
    \begin{equation}
    \mathbf{B}_{\text{fused}} = \left[ \mathbf{B}_{i_1}, \mathbf{B}_{i_2}, \dots, \mathbf{B}_{i_K} \right] \in \mathbb{R}^{k \times rK}.
    \end{equation}
    \begin{equation}
    \Delta\mathbf{W}_{\text{fused}} = \mathbf{B}_{\text{fused}} \mathbf{A}_{\text{fused}}.
    \end{equation}
This fused update $\Delta\mathbf{W}_{\text{fused}}$ integrates knowledge from multiple domains, weighted by their relevance to the test image.

\subsubsection{Updating the Model and Prediction}

We update the model weights for each linear layer:

\begin{equation}
\mathbf{W}' = \mathbf{W} + \Delta\mathbf{W}_{\text{fused}}.
\end{equation}
The adapted model is then used to segment the test image:

\begin{equation}
\hat{\mathbf{y}}_t = \text{CAT-Seg}\left( \Delta \mathcal{W}_{\text{fused}}, \mathbf{x}_t \right).
\end{equation}
This process is executed for each test image individually, enabling real-time, training-free adaptation.

%% file: sec/4_experiments.tex
\input{tables/table_catseg_exclude}

\section{Experiment Setup and Results}
\label{sec:experiments}

To evaluate \methodname{}'s performance in OV semantic segmentation under domain shifts, we design a benchmark with a diverse set of datasets covering substantial data and vocabulary variations. Unlike most of the adaptation literature, often using limited domains or fixed label spaces, our benchmark stress tests the OV capability of the network across varied scenarios, label sets, and weather conditions.

Scalability is essential for practical deployment, so we constructed a large library of LoRA adapters to show that our method scales effectively even with several adapters. This setup mirrors real-world scenarios where models must handle numerous domains and conditions without compromising performance.

\subsection{Datasets}

Our benchmark consists of \textbf{20 diverse semantic segmentation domains}, derived from 10 datasets, selected to represent a wide range of domains and to provide large data and vocabulary shifts. This includes:

\begin{itemize}
    \item \textbf{Driving Datasets}: Commonly used in adaptation scenarios, featuring complex urban scenes with varying conditions -- Cityscapes (CS)~\cite{Cityscapes}, BDD100K (BDD) ~\cite{yu2020bdd100k}, Mapillary Vistas (MV) \cite{neuhold2017mapillary}, and Indian Driving Dataset (IDD) \cite{dokania2023idd}.
    \item \textbf{Weather-Specific Datasets}: Small but challenging datasets focusing on extreme weather conditions like fog, rain, and night-time scenarios -- ACDC~\cite{sakaridis2021acdc}, MUSES~\cite{brodermann2024muses}.
    \item \textbf{General-Purpose Datasets}: Datasets with a vast number of classes, covering a broad spectrum of scenes and objects beyond driving scenarios -- ADE20K~\cite{zhou2017scene} (ADE150), PC59 \cite{mottaghi2014role}, NYU \cite{SilbermanECCV12}, and COCONutL~\cite{coconut2024cvpr}.
\end{itemize}
This diverse collection ensures that our benchmark captures significant variations in both data distribution and label spaces, providing a rigorous evaluation for OV semantic segmentation.

\subsection{Evaluation Protocol}

To adhere to common conventions from the adaptation literature, we adopt a \textbf{leave-one-out} evaluation strategies

\begin{enumerate}
    \item For each dataset, we train an adapter on its training set.
    \item When evaluating on a particular dataset, we \textbf{remove} its corresponding adapter from the library, to prevent the model from leveraging any direct knowledge of it.
    \item We repeat the process for each dataset, ensuring that the model is always tested on unseen data, as it never accesses to domain-specific adapters.
\end{enumerate}

\subsubsection{Implementation Details}

For our \methodname method, we select the top $K=7$ closest LoRA adapters based on CLIP embedding proximity, using a softmax temperature $\tau = 0.01$ to weight the adapters during fusion. This configuration balances performance and computational efficiency. We refer to supplementary material for further details.

\subsection{Comparative Methods}

We compare the following methods:
\begin{itemize}
    \item \textbf{Zero-Shot CAT-Seg \cite{Cho_2024_CVPR}}: The baseline model without any adaptation or LoRA integration.
    \item \textbf{Uniform LoRA Merging}: In this approach, all LoRA adapters (excluding the one corresponding to the test dataset) are uniformly averaged without considering their relevance to the target domain -- which can be seen as an adapters-variant of \cite{li2024training} revised to our setting with supervised fine-tuning of the adapters.
    \item \textbf{\methodname (Ours)}: Our proposed method, which performs adaptive LoRA merging based on domain proximity.
    \item \textbf{Uniform~(\textit{Late Fusion})}: Outputs from all LoRA adapters (excluding the one corresponding to the test dataset) are uniformly averaged at the softmax output level.
    \item \textbf{\methodname~(\textit{Late Fusion})}: Similar to \methodname, weights are computed based on domain proximity but applied to the softmax outputs. This method has higher computational complexity than \methodname{} (one forward pass per adapter).
\end{itemize}
Furthermore, we report the performance achieved by each of the adapters trained on the specific dataset where we evaluate, although such adapters had the unfair advantage of having access to training data from the target domain -- accordingly, we refer to them as \textit{\textbf{Oracles}}. 

\subsection{Main Results and Discussion}

Table~\ref{tab:catseg} presents the mean Intersection over Union (mIoU) scores for each method across all datasets. Our method, \methodname, consistently outperforms both the zero-shot baseline and the uniform merging approach \cite{li2024training} (improvements over the latter are shown in the last row), reaching close to the performance of the single adapters -- and, sometimes, even outperforming them (e.g., on ACDC fog and night). 

One particular dataset, \textit{CoconutL} (labeled with $*$), largely overlaps with the pretraining data of CAT-Seg. Evaluating our performance on it -- even in the leave-one-out protocol -- would not fit the usual setting established for the adaptation task. Therefore, we exclude it from the final harmonic mean calculations.
However, we provide its results in brackets for completeness, which show (unsurprisingly) an exceptional performance of the Zero-shot model.

\textbf{Performance Analysis.}
Our results demonstrate that 
\methodname significantly outperforms the zero-shot baseline across all datasets, highlighting the effectiveness of our domain adaptation strategy.
Uniform merging \cite{li2024training}, despite its simplicity, already shows notable improvements over the zero-shot model, primarily due to linear mode connectivity, showing that LoRA merging is effective when one has access to a LoRA library without any target awareness -- e.g., in domain generalization settings -- as already proven in \citet{li2024training}. 
However, the introduction of careful adapters weighting by \methodname leads to substantial performance gains. This underscores that selecting relevant adapters based on domain proximity is more important than merely scaling up the number of adapters available in the library.
Moreover, we observe that both Uniform and \methodname (\textit{Late Fusion}) perform worse than their LoRA merging counterparts, despite their higher computational complexity. These likely underperform because operating solely at the output level and cannot adjust internal model representations, unlike parameter-level fusion.

\begin{figure}[t]
    \centering
    \includegraphics[width=0.4\textwidth]{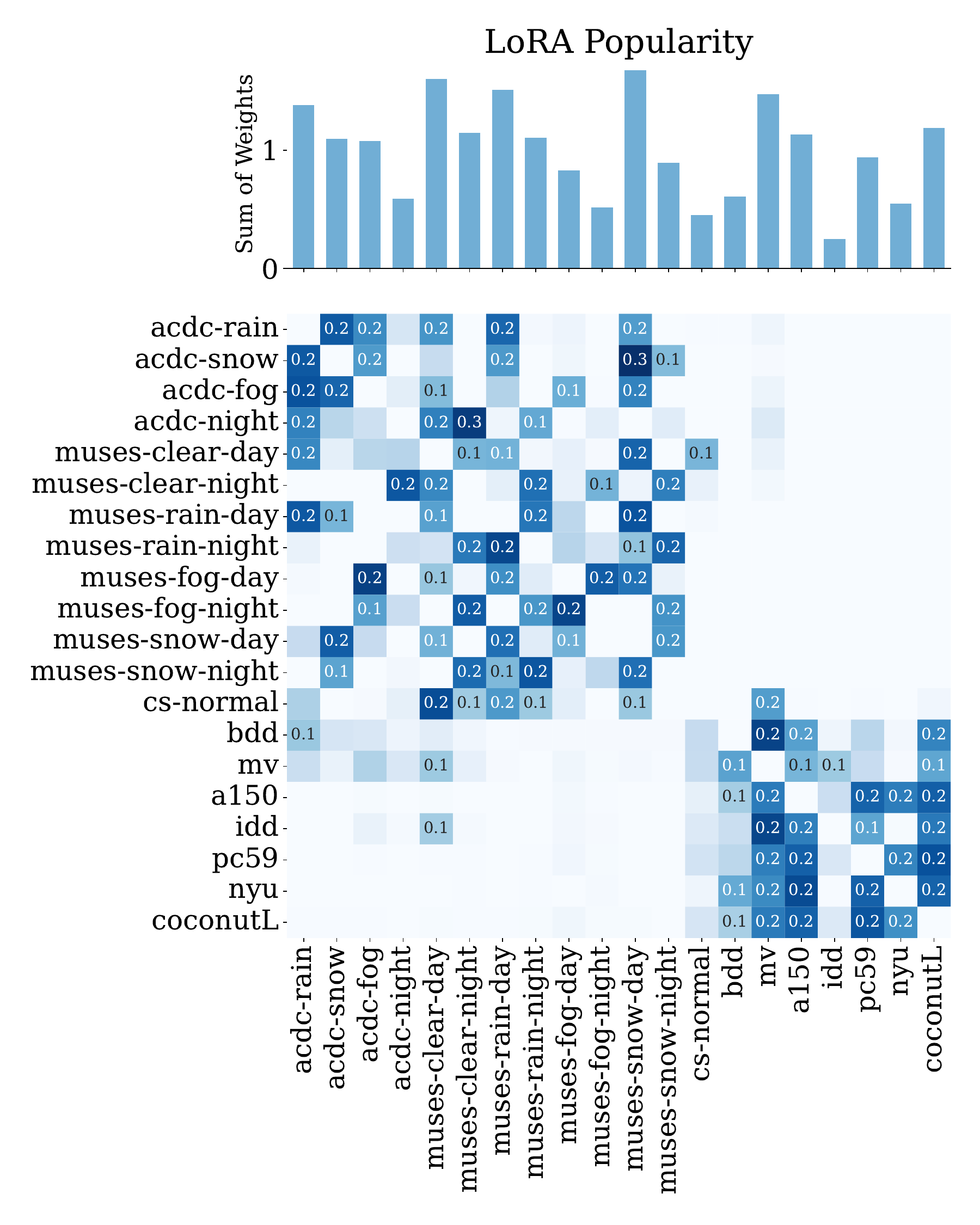}\vspace{-0.3cm}
    \caption{\textbf{Adapter contribution heatmap}. Rows represent individual test datasets, and columns correspond to specific LoRA adapters. The color intensity of each cell indicates the frequency and weight of selection (with values below 0.1 omitted). %
    The diagonal is empty due to the leave-one-out strategy. }\vspace{-0.3cm}
    \label{fig:adapter_contribution}
\end{figure}

\input{tables/table_catseg_include}

\textbf{Adapter Contribution Analysis.}
To understand how the adapters contribute across datasets, we analyze the selection patterns shown in Figure~\ref{fig:adapter_contribution}, where a heatmap indicates the frequency and weight of each adapter's selection per test dataset. Adapter support is spread across the library, showing that the model benefits from multiple adapters rather than relying on a few only. 
On the one hand, we can appreciate high weights for strongly related domains -- e.g., MUSES-snow-day heavily relies on ACDC-snow and vice-versa; in general, we can observe polarized selection for driving datasets (top left) or generic ones (bottom right).
Nonetheless, even adapters from unrelated domains are often selected, suggesting the existence of shared useful features, though each category still predominantly influences related datasets. By examining columns (and aggregating them in a histogram), certain adapters emerge as ``popular" choices, highlighting their stronger influence.

\begin{figure}[t]
    \centering    \includegraphics[width=0.42\textwidth]{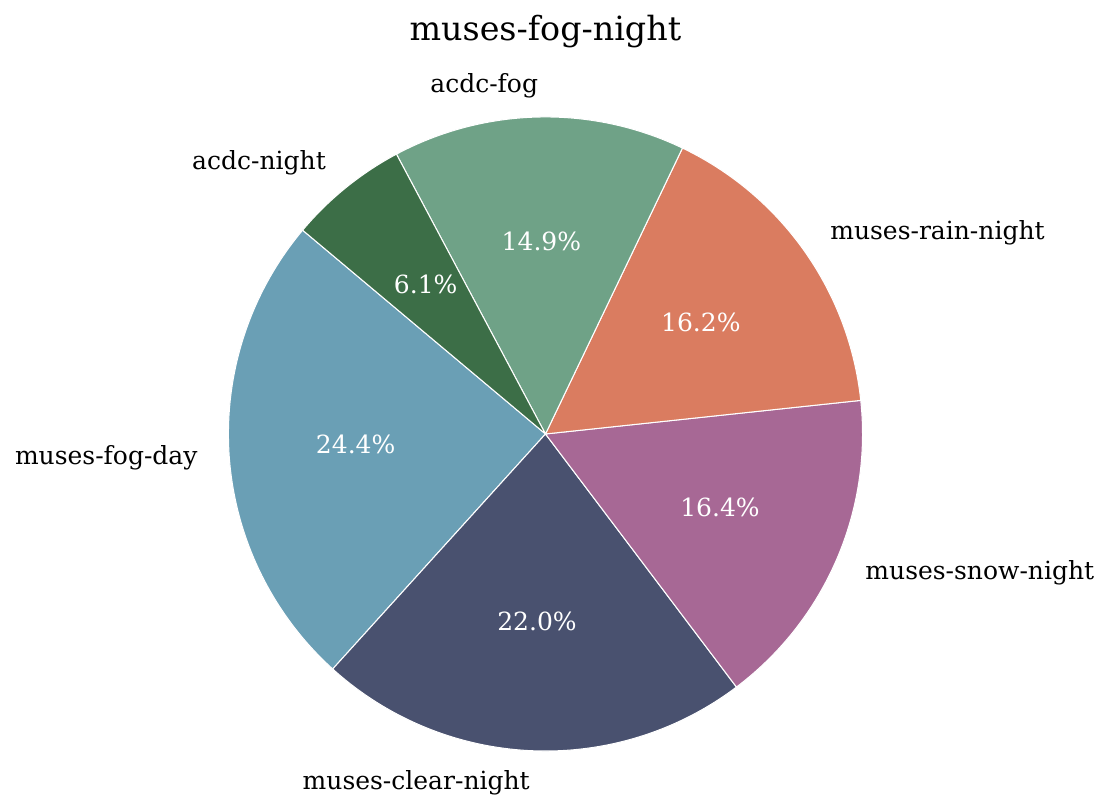}\vspace{-0.3cm}
    \caption{\textbf{Adapter weight distribution for MUSES-Fog-Night.} The fused adapter combines knowledge from foggy and night-time conditions by weighting relevant adapters. Adapters with a weight lower than 5\% are not included.}\vspace{-0.3cm}
    \label{fig:piechart}
\end{figure}

\subsection{Analysing CLIP as a Domain Navigator}

A key assumption of our framework is that CLIP embeddings provide a meaningful mapping of the domains and that proximity in the CLIP space correlates with segmentation performance. To verify this assumption, we examine i) how our method navigates complex domains by combining relevant adapters; and ii) whether proximity in the CLIP space is a good heuristic for predicting performance.

\textbf{Domain Composition Analysis.} 
We observe that CLIP effectively navigates the domain space, as shown in the adapter selection for \textbf{MUSES-Fog-Night}. This dataset, comprising foggy night images, lacks a direct match in our library. However, adapters trained on night and fog conditions are available.
Examining LoRA contributions for MUSES-Fog-Night in Figure~\ref{fig:piechart}, we see that fog and night adapters dominate with around 40$\%$ and 60$\%$ contribution respectively, with higher weights for MUSES adapters, around 80$\%$, likely due to shared camera characteristics. This illustrates our ability to combine relevant knowledge even when an exact match is absent and showcases the easy interpretability of our approach.

\begin{figure}[t]
    \centering
    \includegraphics[width=0.45\textwidth]{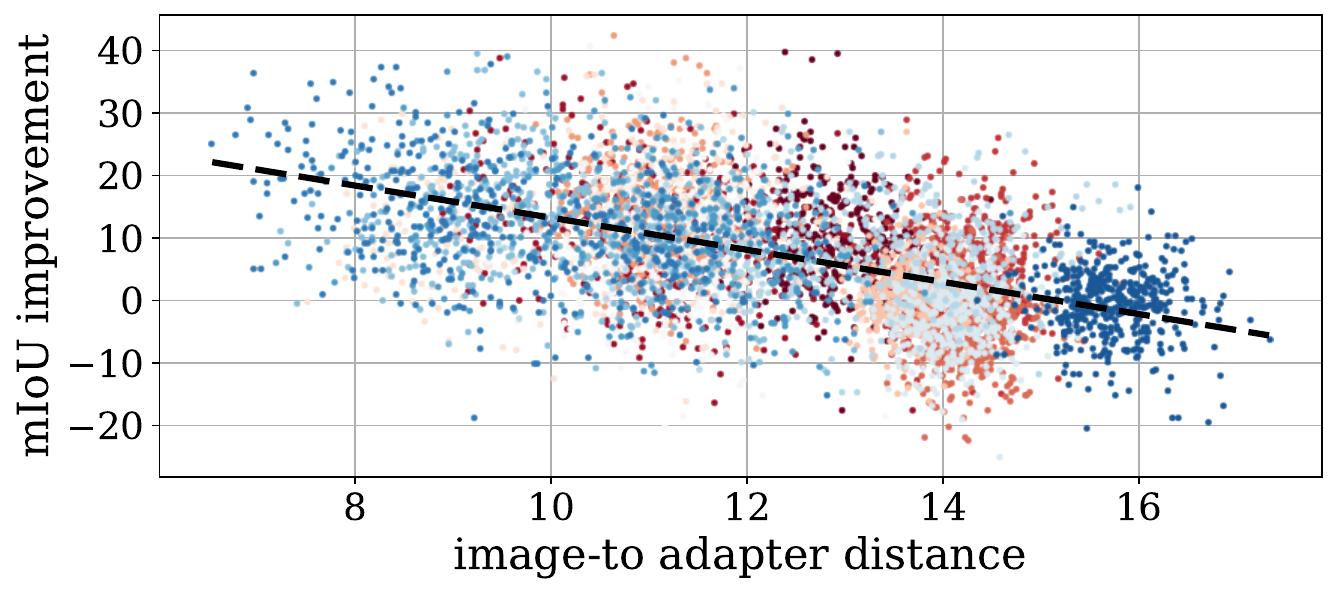}
    \caption{\textbf{CLIP-guidance effectiveness for LoRA selection on ACDC.} Each point represents an image-adapter combination, with adapters separated by color. x-axis: distance from the corresponding image embedding to the adapter embedding. y-axis: improvement in mIoU when using the adapter relative to the zero-shot base network. The linear regression curve (dashed line) indicates that embedding similarity correlates with higher mIoU. 
    We show the full adapter library, excluding those trained on ACDC.
    }
    \label{fig:correlation}
\end{figure}

\textbf{CLIP Distance - Performance Correlation.}
Having observed that CLIP embeddings effectively guide adapter selection, we proceed to the second question: \emph{Is image-LoRA proximity in the CLIP space a good heuristic for predicting segmentation performance?}
Figure~\ref{fig:correlation} indicates that this is indeed the case.
The negative slope of the regression line indicates an inverse relationship between CLIP distance and mIoU -- 
i.e., smaller CLIP distances (higher proximity) generally correspond to better segmentation. This empirical observation confirms that CLIP embedding distance is a good heuristic for predicting adapter effectiveness.

\input{tables/table_sed_exclude}

\begin{figure*}[t]
    \centering
    \includegraphics[trim=0.5cm 0cm 0.5cm 0cm, clip,width=0.95\textwidth]{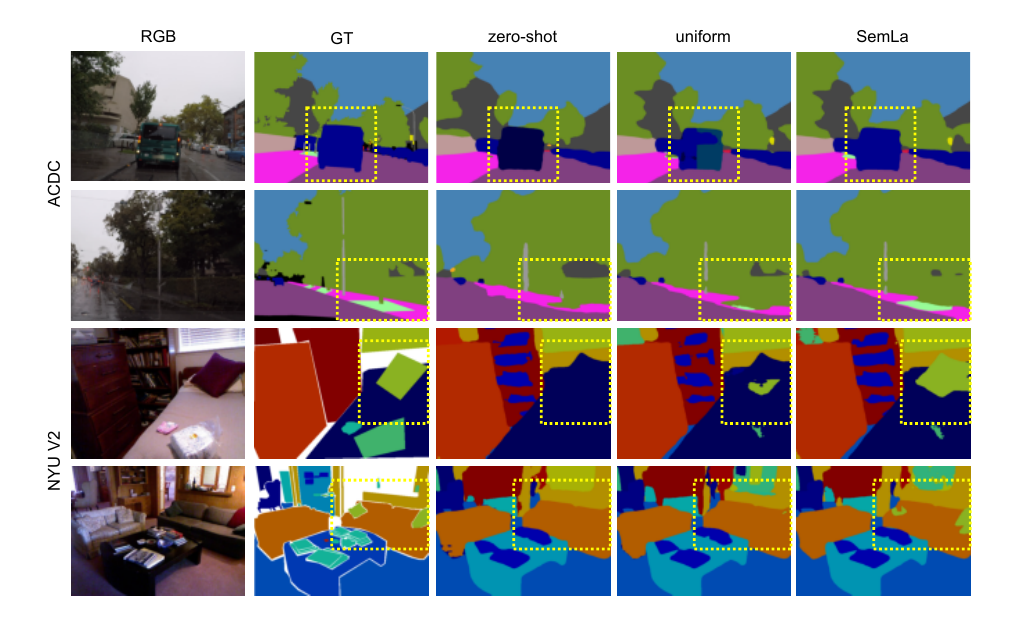}\vspace{-0.75cm}
    \caption{\textbf{Qualitative results on ACDC and NYU.} From left to right: Input Image (RGB), ground-truth semantic masks (GT) and predictions by the Zero-Shot model \cite{Cho_2024_CVPR}, Uniform Merging \cite{li2024training}, and \methodname.}\vspace{-0.4cm}
    \label{fig:qualitative}
\end{figure*}

\subsection{All-Inclusive Settings}
We evaluate the -- unlikely -- scenario where all the LoRA adapters corresponding to the benchmark datasets are included in the library at test-time (i.e., in contrast with common adaptation settings). 
Table \ref{tab:catseg_allinclusive} reports \methodname tested with two $\tau$ and Uniform~\cite{li2024training} settings. The results reveal a two-fold effect: when the system is directed to prioritize the target-domain LoRA adapter, the model's performance almost matches the Oracle. However, when the temperature $\tau$ is higher, hence LoRA fusion is promoted, the model -- while slightly lower on average -- overcomes the Oracles on 14 domains over 20, primarily in highly related datasets. This finding underscores that LoRA synergies can extend beyond individual adapters, achieving performance that, in some cases, surpasses even the Oracle.

\subsection{Results with a Different Backbone: SED}
Our method is inherently backbone-agnostic, as we only operate on LoRA retrieval and adaptation, relying on simple and widespread methods. To empirically prove this claim, we generate another library based on \textbf{SED}~\cite{xie2024sed}, another popular open-vocabulary semantic segmentation model. 
In Table \ref{tab:sed}, we collect the outcome of this new evaluation, carried out once again on our 20-domain benchmark. All our achievements with CAT-Seg are confirmed when switching to SED, with \methodname clearly outperforming both the zero-shot model and the uniform merging strategy \cite{li2024training}, thus confirming the flexibility of our framework and the possibility of tailoring it around different OV semantic segmentation backbones.

\subsection{Qualitative Results}
Figure~\ref{fig:qualitative} presents qualitative comparisons between the predictions by different methods -- zero-shot CAT-Seg, uniform merged model \cite{li2024training} and ours. \methodname produces more accurate and detailed segmentation maks, demonstrating its superiority over zero-shot and uniform merging models.

\subsection{Limitations and Future Work}

While our method demonstrates strong performance, several areas remain for future exploration. First, we do not explicitly address vocabulary alignment between source and target domains; incorporating vocabulary matching could enhance performance by ensuring compatibility with the target label space. Second, exploring automated domain discovery through CLIP clustering and unsupervised training of new LoRA adapters could lead to a self-augmenting library, enhancing adaptability without manual intervention. Scaling \methodname to thousands of adapters introduces challenges in recognizing and addressing library weaknesses and gaps; developing automated strategies to identify and mitigate them will be essential. Lastly, applying our approach to multi-tasks settings, \textit{e.g.} panoptic segmentation or depth estimation \cite{Poggi_2024_CVPR}, could validate its versatility.

%% file: tables/table_catseg_exclude.tex
\begin{table*}[t]
    \centering
    \setlength{\tabcolsep}{3.5pt}
    \scalebox{0.6}{
    \begin{tabular}{l|cccc|cccccccc|cccccccc|c}
    \hline
    \multirow{2}{*}{Method}
      & \multicolumn{4}{c|}{ACDC}
      & \multicolumn{8}{c|}{MUSES}
      & \multirow{2}{*}{CS}
      & \multirow{2}{*}{BDD}
      & \multirow{2}{*}{MV}
      & \multirow{2}{*}{A150}
      & \multirow{2}{*}{IDD}
      & \multirow{2}{*}{PC59}
      & \multirow{2}{*}{NYU}
      & \multirow{2}{*}{COCONutL*}
      & \multirow{2}{*}{h-mean} \\
    & rain & snow & fog & night
    & clear (d) & clear (n) & rain (d) & rain (n) & fog (d) & fog (n) & snow (d) & snow (n)
    & & & & & & & & \\
    \hline
    Zero-shot~\cite{Cho_2024_CVPR}
      & 46.53 & 48.04 & 47.09 & 37.93
      & 44.43 & 39.29 & 38.95 &27.78 & 53.73 & 25.35 & 43.56 & 33.29 & 47.11 & 47.95
      & 25.69 & 37.83 & 35.39 & 63.33 & 49.38 & (68.26)
      & 39.39 \\
      
    Oracles
      & 70.94 & 69.22 & 69.98 & 51.55 & 69.36 & 57.09 & 54.28 & 52.11 & 75.85 & 61.26 & 66.25 & 54.35 & 67.47 & 60.06 & 49.57 & 53.99 & 64.34 & 68.68 & 61.90 & (70.44) & 61.05 \\
    \hline
    Uniform (\textit{Late Fusion}) 
        & 64.58 & 64.76 & 69.64 & 48.51 & 60.53 & 51.88 & 50.62 & 37.30 & 79.00 & 35.68 & 62.39 & 44.04 & 61.82 & 57.08 & 30.34 & 36.04 & 37.43 & 63.24 & 47.39 & (66.40) & 49.26 \\

    \methodname (\textit{Late Fusion}) 
        & 66.09 & 66.77 & 71.01 & 50.79 & 62.69 & 55.70 & 53.27 & 45.71 & 68.86 & 45.65 & 65.20 & 49.71 & 62.86 & 57.60 & 29.68 & 37.28 & 39.64 & 63.87 & 50.64 & (65.84) & 52.09  \\
    \hline
    Uniform~\cite{li2024training}
      & 67.40 & 66.35 & 69.71 & 49.98 & 58.28 & 55.78 & 54.70 & 45.09 & \textbf{73.75} & 45.16 & 61.02 & 49.08 & 62.18 & \textbf{58.19} & \textbf{31.51} & 37.25 & 38.83 & 63.06 & 48.93 & (\textbf{67.62}) & 51.89 \\

    \bf \methodname{} (ours)
      & \textbf{67.71} & \textbf{68.95} & \textbf{71.92} & \textbf{51.73} & \textbf{61.09} & \textbf{60.06} & \textbf{57.60} & \textbf{47.35} & 72.97 & \textbf{52.38} & \textbf{67.28} & \textbf{55.92} & \textbf{63.91} & {57.30} & 31.12 & \textbf{38.18} & \textbf{40.16} & \textbf{64.75} & \textbf{51.35} & (67.26) & \textbf{54.16} \\
    & \textcolor{forestgreen}{+0.31} & \textcolor{forestgreen}{+2.60} & \textcolor{forestgreen}{+2.21} & \textcolor{forestgreen}{+1.75} & \textcolor{forestgreen}{+2.81} & \textcolor{forestgreen}{+4.28} & \textcolor{forestgreen}{+2.90} & \textcolor{forestgreen}{+2.26} & \textcolor{burgundy}{-0.78} & \textcolor{forestgreen}{+7.22} & \textcolor{forestgreen}{+6.26} & \textcolor{forestgreen}{+6.84} & \textcolor{forestgreen}{+1.73} & \textcolor{burgundy}{-0.89} & \textcolor{burgundy}{-0.39} & \textcolor{forestgreen}{+0.93} & \textcolor{forestgreen}{+1.33} & \textcolor{forestgreen}{+1.69} & \textcolor{forestgreen}{+2.42} & \textcolor{burgundy}{-0.36} & \textcolor{forestgreen}{+2.27} 
     
     \\
    \hline
    \end{tabular}}
    \vspace{-0.3cm}
    \caption{\textbf{Adaptation for OV semantic segmentation -- CAT-Seg \cite{Cho_2024_CVPR}.}  
    Performance comparison across our 20-domain benchmark, leave-one-out setting. On MUSES, (d) and (n) stand for \textit{day} and \textit{night}. ( ) means excluded from h-mean. \methodname{} with $\tau=0.05$, and $K=5$.} \vspace{-0.3cm}
    \label{tab:catseg}
\end{table*}

%% file: tables/table_catseg_include.tex
\begin{table*}[t]
    \centering
    \setlength{\tabcolsep}{3.3pt}
    \scalebox{0.62}{
    \begin{tabular}{l|cccc|cccccccc|cccccccc|c}
    \hline
    \multirow{2}{*}{Method}
      & \multicolumn{4}{c|}{ACDC}
      & \multicolumn{8}{c|}{MUSES}
      & \multirow{2}{*}{CS}
      & \multirow{2}{*}{BDD}
      & \multirow{2}{*}{MV}
      & \multirow{2}{*}{A150}
      & \multirow{2}{*}{IDD}
      & \multirow{2}{*}{PC59}
      & \multirow{2}{*}{NYU}
      & \multirow{2}{*}{COCONutL*}
      & \multirow{2}{*}{h-mean} \\
    & rain & snow & fog & night
    & clear (d) & clear (n) & rain (d) & rain (n) & fog (d) & fog (n) & snow (d) & snow (n)
    & & & & & & & & \\
    \hline

    Oracles
      & 70.94 & 69.22 & 69.98 & 51.55 & 69.36 & 57.09 & 54.28 & 52.11 & 75.85 & 61.26 & 66.25 & 54.35 & 67.47 & 60.06 & \textbf{49.57} & \textbf{53.99} & \textbf{64.34} & \textbf{68.68} & 61.90 & (\textbf{70.44}) & 61.05 \\
    \hline

   Uniform
      & 67.76 & 66.69 & 69.91 & 50.44 & 58.60 & 56.30 & 55.73 & 45.96 & 73.33 & 46.29 & 61.36 & 50.27 & 62.77 & 58.42 & 32.87 & 38.48 & 39.89 & 63.77 & 50.12 & (68.13) & 52.78 
     
     \\

          \methodname{} $\tau=0.005$
      & \textbf{70.97} & {69.40} & {70.94} & {51.68} & \textbf{69.51} & 57.95 & 54.25 & 52.14 &\textbf{ 75.88} & 61.26 & 66.16 & 51.22 & \textbf{67.52} & \textbf{60.20} & 48.70 & 49.42 & 64.27 & 68.27 & \textbf{62.68} & (68.83) & 60.57 \\

          \methodname{} $\tau=0.05$
      & 68.82 & \textbf{70.43} & \textbf{73.38} &\textbf{53.06} & 67.19 & \textbf{59.93} & \textbf{59.30} & \textbf{52.21} & 72.45 & \textbf{62.10} & \textbf{69.33} & \textbf{59.38} & 67.49 & 58.18 & 38.06 & 44.20 & 49.49 & 67.45 & 57.93 & (68.96) & 58.79 \\
    \hline
    \end{tabular}}
    \vspace{-0.3cm}
    \caption{
    \textbf{Adaptation for OV semantic segmentation -- CAT-Seg \cite{Cho_2024_CVPR}, all-inclusive setting.}
    Performance comparison across our 20-domain benchmark, 
    when \textit{all} domains are available in the LoRA Library, hence to \textit{source}. On MUSES, (d) and (n) stand for \textit{day} and \textit{night}. ( ) means excluded from h-mean. \methodname{} with $K=5$.}
    \label{tab:catseg_allinclusive}\vspace{-0.3cm}
\end{table*}

%% file: tables/table_sed_exclude.tex
\begin{table*}[t]
    \centering
    \setlength{\tabcolsep}{3.5pt}
    \scalebox{0.62}{
    \begin{tabular}{l|cccc|cccccccc|cccccccc|c}
    \hline
    \multirow{2}{*}{Method}
      & \multicolumn{4}{c|}{ACDC}
      & \multicolumn{8}{c|}{MUSES}
      & \multirow{2}{*}{CS}
      & \multirow{2}{*}{BDD}
      & \multirow{2}{*}{MV}
      & \multirow{2}{*}{A150}
      & \multirow{2}{*}{IDD}
      & \multirow{2}{*}{PC59}
      & \multirow{2}{*}{NYU}
      & \multirow{2}{*}{COCONutL*}
      & \multirow{2}{*}{h-mean} \\
    & rain & snow & fog & night
    & clear (d) & clear (n) & rain (d) & rain (n) & fog (d) & fog (n) & snow (d) & snow (n)
    & & & & & & & & \\
    \hline
    Zero-shot~\cite{Cho_2024_CVPR}
      & 42.64 & 43.97 & 46.88 & 35.06 & 38.05 & 35.01 & 35.18 & 27.87 & 48.09 & 19.53 & 42.63 & 23.53 & 41.90 & 43.41 & 25.05 & 35.27 & 35.20 & 60.87 & 47.20 & (66.36) & 35.56 \\
    Oracles
      & 61.70 & 68.25 & 69.33 & 48.39 & 62.04 & 55.55 & 53.61 & 44.82 & 69.40 & 53.20 & 65.60 & 55.85 & 68.48 & 58.61 & 47.28 & 50.66 & 63.15 & 66.55 & 60.60 & (68.01) & 58.05 \\
    \hline
    Uniform~\cite{li2024training}
      & 59.59 & 61.07 & 68.33 & 47.27 & 58.79 & 53.87 & 50.65 & 37.12 & \bf 69.82 & 35.52 & 61.46 & 44.08 & 61.85 & \bf 54.78 & \bf 30.37 & 34.92 & 39.59 & 60.82 & 48.42 & (64.05) & 48.60 \\
    \bf \methodname{} (ours)
      & \bf 60.01 & \bf 67.00 & \bf 69.96 & \bf 49.70 & \bf 60.98 & \bf 54.79 & \bf 55.15 & \bf 38.89 & 69.46 & \bf 38.25 & \bf 66.80 & \bf 48.89 & \bf 64.35 & 54.40 & 28.99 & \bf 36.91 & \bf 41.08 & \bf 62.23 & \bf 51.82 & \bf (65.46) & \bf 50.57 \\
      & \textcolor{forestgreen}{+0.42} & \textcolor{forestgreen}{+5.93} & \textcolor{forestgreen}{+1.63} & \textcolor{forestgreen}{+2.43} & \textcolor{forestgreen}{+2.19} & \textcolor{forestgreen}{+0.92} & \textcolor{forestgreen}{+4.50} & \textcolor{forestgreen}{+1.77} & \textcolor{burgundy}{-0.38} & \textcolor{forestgreen}{+2.70} & \textcolor{forestgreen}{+5.34} & \textcolor{forestgreen}{+4.81} & \textcolor{forestgreen}{+2.50} & \textcolor{burgundy}{-0.38} & \textcolor{burgundy}{-1.38} & \textcolor{forestgreen}{+1.99} & \textcolor{forestgreen}{+1.59} & \textcolor{forestgreen}{+1.69} & \textcolor{forestgreen}{+3.40} & \textcolor{forestgreen}{+1.41} & \textcolor{forestgreen}{+1.97} \\
    \hline
    \end{tabular}}
    \vspace{-0.3cm}
    \caption{
    \textbf{Adaptation for OV semantic segmentation -- SED \cite{xie2024sed}.}
    Performance comparison across 20-domain benchmark, leave-one-out setting. 
    On MUSES, (d) and (n) stand for day and night. ( ) means excluded from h-mean. SemLA with $\tau=0.05$, and
$K=5$.}\vspace{-0.7cm}
    \label{tab:sed}
\end{table*}

%% file: sec/5_conclusions.tex
\section{Conclusion}
\label{sec:conclusion}
We have presented \methodname, a framework for training-free test-time adaptation in open-vocabulary semantic segmentation. By constructing a library of LoRA adapters indexed with CLIP embeddings and dynamically merging them based on domain proximity, our method effectively mitigates domain shifts.
Extensive experiments show that \methodname outperforms zero-shot baselines, uniform adapter merging, and sometimes even fine-tuned models across diverse datasets. We confirmed that CLIP embeddings reliably guide adapters selection and that proximity in the embedding space is a good heuristic for performance.
With enhanced explainability, scalability, and data privacy, \methodname is suitable for practical applications. Future work will address vocabulary alignment, automated domain discovery, and extending the framework to other tasks.

\small{\textbf{Ackonwledgments.} The authors thank Hedvig Kjellstr{\"o}m for the helpful discussions and guidance, and acknowledge The European High Performance Computing Joint Undertaking (EuroHPC JU), EuroCC National Competence Center Sweden (ENCCS) and the CINECA award under the ISCRA initiative for the availability of high-performance computing resources and support.

This study was funded by the European Union – Next Generation EU within the framework of the National Recovery and Resilience Plan NRRP – Mission 4 ``Education and Research" – Component 2 - Investment 1.1 ``National Research Program and Projects of Significant National Interest Fund (PRIN)" (Call D.D. MUR n. 104/2022) – PRIN2022 – Project reference: ``RiverWatch: a citizen-science approach to river pollution monitoring" (ID: 2022MMBA8X, CUP: J53D23002260006).}

%% file: sec/X_suppl.tex
\setcounter{page}{1}

\twocolumn[{%
\renewcommand\twocolumn[1][]{#1}%
\maketitlesupplementary
\centering
    \includegraphics[width=0.98\textwidth, trim=2cm 1.7cm 0 0, clip]{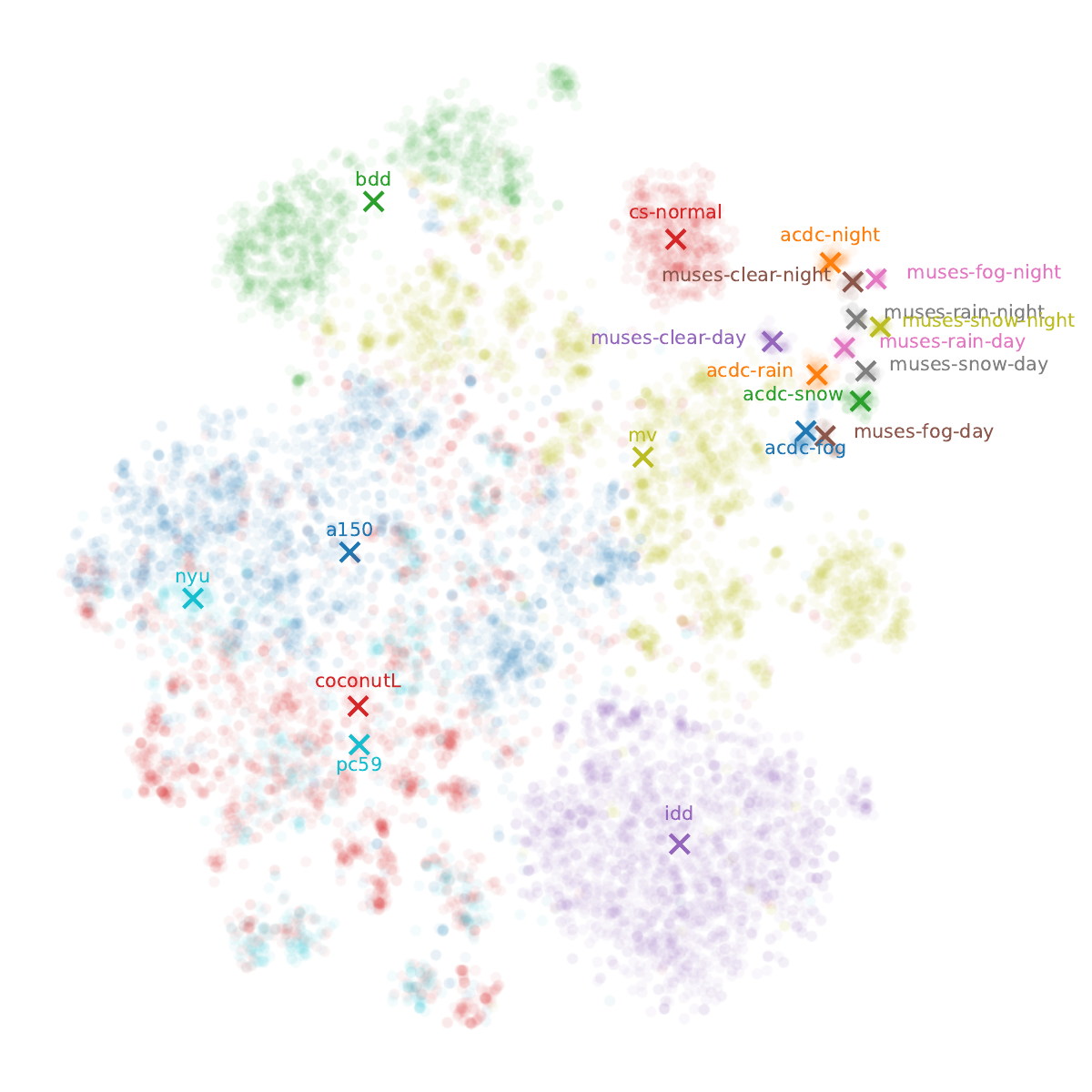}
   \captionof{figure}{\textbf{t-SNE visualization of the LoRA Library and relative datasets in the CLIP embedding space.} Similar domains are clustered together, indicating areas with higher LoRA support and potentially stronger performance improvements. A maximum of 15,000 samples per dataset is used for the t-SNE fit and only 15\% of the total datapoints are used for plotting.}\label{fig:tsne_lora_library}
   \vspace{2cm}
  \label{fig:teaser}  
}]

\appendix

\input{tables/table_blip}

\begin{figure*}[t]
    \centering
\includegraphics[width=\textwidth]{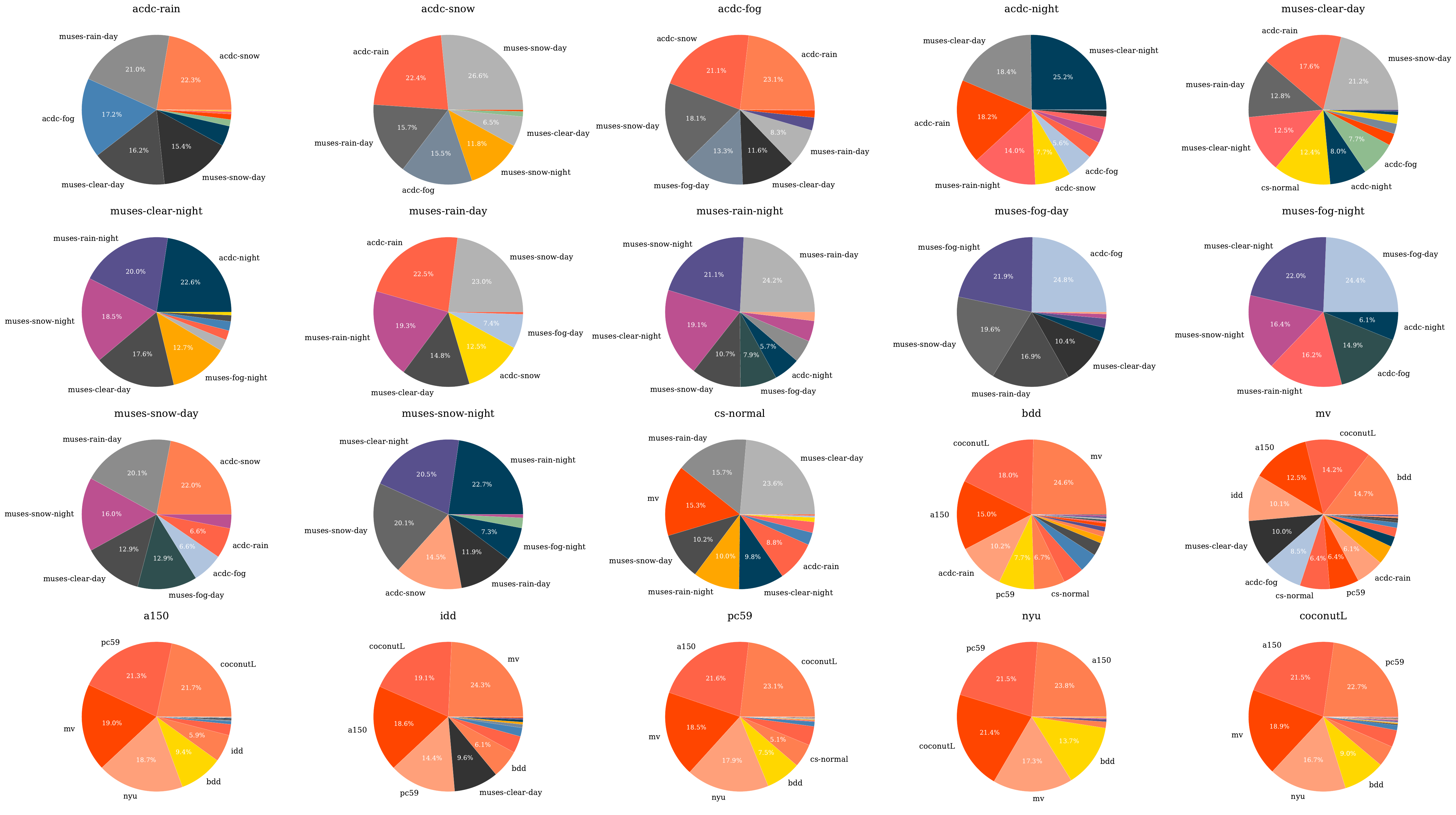}\vspace{-0.3cm}
        \caption{\textbf{Adapters weight distribution for each benchmark dataset.} Each pie chart is divided into sections proportional to the average contribution provided by each adapter based on CAT-Seg leave-one-out adaptation settings.}\vspace{-0.3cm}
    \label{fig:piechart_example}
\end{figure*}

\section*{Introduction}

In this supplementary document, we provide additional details and analyses to support and extend the findings presented in the main paper. Here we report additional details about the implementation specifics, auxiliary experimental results, extensive ablation studies, discuss practical considerations for real-world deployment, and offer additional qualitative examples that highlight the effectiveness and robustness of our proposed method, \methodname. The rest of this document is organized as follows:

\begin{itemize}
    \item \textbf{Section~\ref{sec:implementation_and_code}} details the implementation aspects of our method, including training procedures, hyperparameters, model architecture modifications, and code availability for replication purposes.
    \item \textbf{Section~\ref{sec:data_and_library_analysis}} provides an in-depth analysis of the LoRA adapter library, including visualizations of the embeddings from training samples using t-SNE, labeling of the adapters with natural language using BLIP-2~\cite{li2023blip} for the sake of interpretability, analyses of adapter contributions dataset by dataset, and support score analysis.
    \item \textbf{Section~\ref{sec:ablations}} presents extensive ablation studies and performance analyses. We %
    explore the effectiveness of fully fine-tuned models versus LoRA adapters, conduct hyperparameter sensitivity analysis, and assess alternative domain navigators such as DINOv2~\cite{oquabdinov2} versus CLIP.
    \item \textbf{Section~\ref{sec:additional_qualitative_results}} showcases additional qualitative results across various domains, further demonstrating the adaptability and efficacy of our approach.
    \item \textbf{Section~\ref{sec:discussion}} discusses pragmatic considerations for real-world deployment of \methodname, including strategies to handle computational overhead, domain navigation in specialized domains, scalability concerns, and methods to ensure efficiency and reliability in production environments.
\end{itemize}

\section{Implementation Details}
\label{sec:implementation_and_code}

\paragraph{LoRA Training Details.}
\label{sec:lora_training_details}
We attach LoRAs to every \textit{nn.Linear} layer in the CAT-Seg architecture, except for CLIPs token embedding layer, as this parameter was not trained in the original CAT-Seg implementation either. All LoRAs are trained with the same LoRA configuration -- i.e., rank $r$=8 and $\alpha$=16. The training hyperparameters are largely the same as CATSeg with minor modifications: compared to the original CAT-Seg implementation, we use a base learning rate of 1e-4, weight decay of 1e-5, and 1000 warm-up iterations. For ACDC and MUSES adapters we use a batch size of 2 and a warm-up factor of 0.01. For BDD and CS we increase the batch size to 4 while keeping other hyperparameters the same. For the remaining datasets, we increased the warm-up factor to 0.1 while keeping other hyperparameters the same. All the adapters were trained until convergence.

\vspace{-0.3cm}\paragraph{Code and Models.}
\label{sec:code_and_models}
Full source code and documentation are available in our project page \projectpage.

\section{Interpretability of the LoRA Library}
\label{sec:data_and_library_analysis}

\subsection{t-SNE Visualization of the LoRA Library}
\label{sec:lora_library_visualization}

Figure~\ref{fig:tsne_lora_library} presents a t-SNE~\cite{tsnepaper} visualization of the LoRA adapters' centroids and their associated datasets in the CLIP embedding space. This visualization illustrates the distribution and relationships among different adapters and domains, highlighting similarities between them.

We observe that domains with similar visual characteristics are positioned closely, such as foggy conditions or nighttime scenes. This clustering validates the effectiveness of using CLIP embeddings for adapter selection.

\subsection{BLIP for Labeling LoRA Adapters}
\label{sec:blip_lora_labeling}
To highlight the transparency and interpretability of our system, we leverage the connection between the CLIP embedding space and natural language. By processing the centroids of our LoRA adapters with BLIP-2~\cite{li2023blip}, we obtain natural language captions describing the domain encapsulated in each adapter's training set. This provides semantic insights into the content and characteristics of the training set used for each adapter. 

Table \ref{tab:blip2} reports, for each dataset, the caption provided by BLIP, together with the answers to two simple questions ``Where is this?" and ``Describe the environment in two words", when processing domain centroids. We can appreciate how both captions and answers are strongly related to the content in each dataset, even though retrieved information remains limited and coarse. Nevertheless, the possibility of extracting natural language captions of the LoRA centroids is an interesting feature, further motivating the use of CLIP as our domain navigator.

\subsection{Adapter Contributions}
\label{sec:adapter_piecharts}

Figure~\ref{fig:piechart_example} shows the adapter weight distribution for all datasets composing our benchmark involved in the leave-one-out experiments. The parameters used in this experiment are $\tau=0.01$ and top-${K}=7$. The weights represent the relative contribution of each adapter to the fused model, highlighting their respective roles in the overall composition.

These pie charts provide insights into how different adapters contribute to the final model for specific target domains, demonstrating the effective combination of knowledge from relevant adapters.

\subsection{LoRA Support Score Analysis}
\label{sec:lora_support_analysis}

We analyze the relationship between the LoRA support score and mIoU performance. We define LoRA support score for a test image $\mathbf{x}_t$:

\begin{equation}
\text{Support Score}(\mathbf{x}_t) = \sum_{i \in \mathcal{K}} \frac{w_i}{\left\| \mathbf{e}_t - \mathbf{c}_i \right\|_2},
\end{equation}

where $w_i$ is the weight assigned to adapter $i$, $\mathbf{e}_t$ is the CLIP embedding of the test image, $\mathbf{c}_i$ is the centroid of adapter $i$, and $\mathcal{K}$ is the set of top-$K$ selected adapters.

\begin{figure*}[t]
    \centering
    \includegraphics[width=\textwidth]{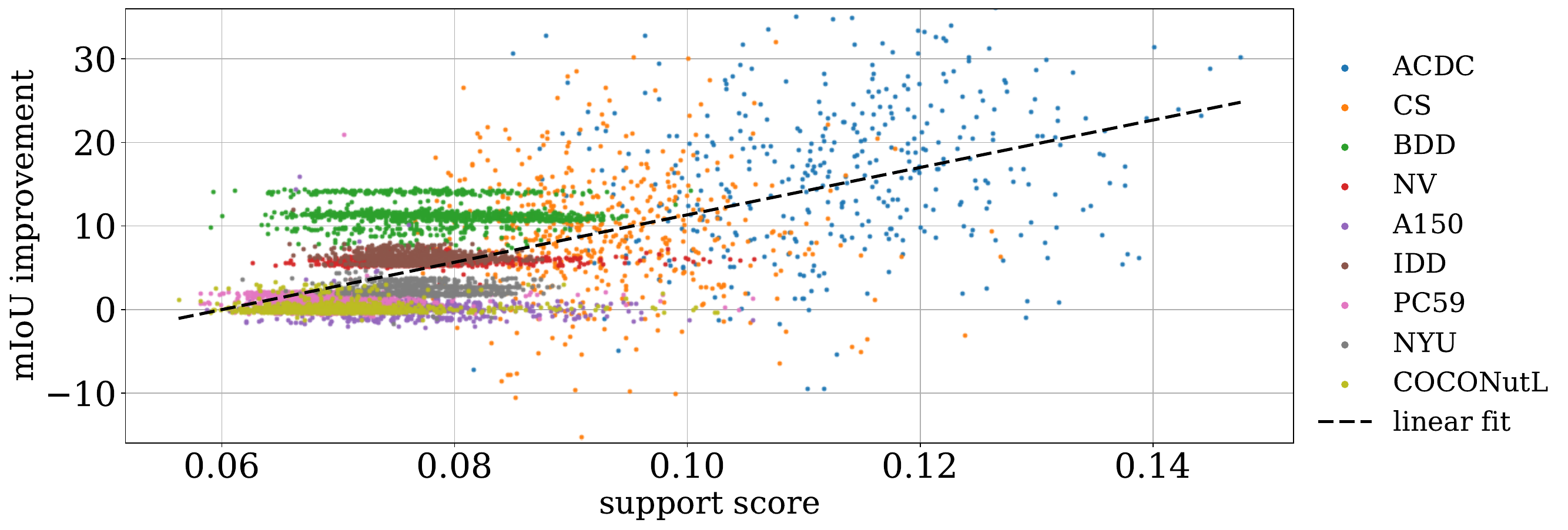}
    \caption{\textbf{Correlation between LoRA support score and mIoU.} Higher support scores correlate with better segmentation performance.}
    \label{fig:support_score_analysis}
\end{figure*}

We compute the support score for a sample of images and plot it against their corresponding mIoU scores. Figure~\ref{fig:support_score_analysis} shows that images with higher support scores tend to have higher mIoU, confirming that the LoRA support score is a good predictor of segmentation performance. We also notice that for low values of support score (\textit{e.g.} below 0.09), an improvement in support score does not strictly imply a stronger improvement in mIoU, showing that the underlying relation is likely not linear.

Overall, this analysis validates our assumption that proximity in the CLIP embedding space, combined with the weighting mechanism, is an effective heuristic for adapter selection.

\begin{figure}[t]
    \centering
    \includegraphics[width=0.5\textwidth]{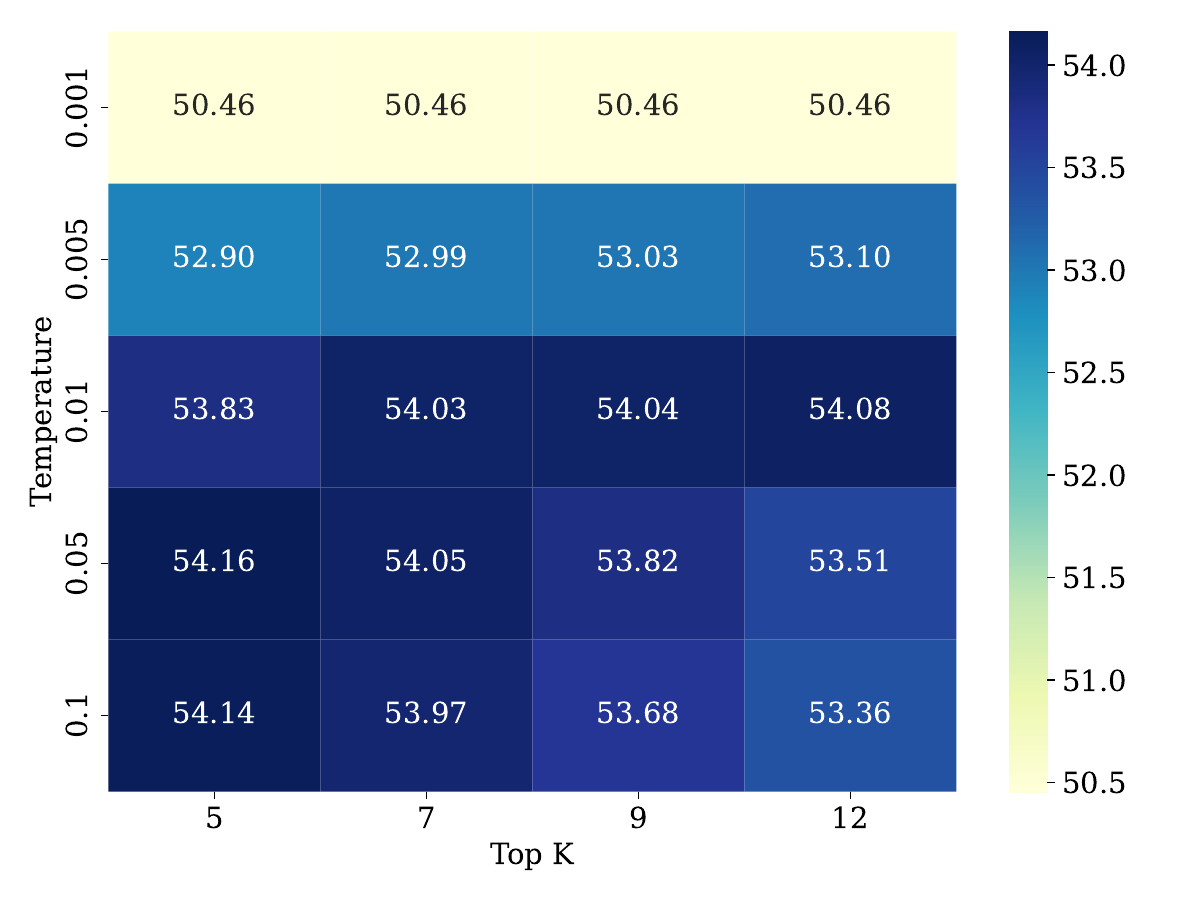}\vspace{-0.4cm}
    \caption{\textbf{Hyper-parameters Study.} Impact of $K$ (number of adapters) and $\tau$ (temperature) on overall performance (mIoU).}
    \label{fig:heatmap_hyperparameters}
\end{figure}

\input{tables/distance_ablations}

\input{tables/table_fft}

\input{tables/table_dino.tex}

\section{Ablations and Analysis}
\label{sec:ablations}

\subsection{Hyper-parameters Study}
\label{sec:hyperparameter_analysis}

We conduct ablations over $\tau$ and $K$, the two hyper-parameters controlling our system at test time.

\begin{itemize}
    \item \textbf{Number of Adapters ($K$)}: Increasing $K$ includes more adapters in the fusion, potentially providing more context but risking the introduction of unrelated knowledge while increasing the LoRA merging computational overhead.
    \item \textbf{Temperature ($\tau$)}: Regulates the weighting of adapters based on their distances. Lower $\tau$ emphasizes closer adapters; higher $\tau$ promotes a more uniform weighting.
\end{itemize}

Figure~\ref{fig:heatmap_hyperparameters} shows a heatmap of overall performance across different values of $K$ and $\tau$.
Performance peaks at $K=7$ and $\tau=0.01$, balancing relevance and diversity in adapter selection.

\subsection{Distance Metrics Comparison}
\label{sec:distance_comparison}

We compare Euclidean distance (used in \methodname) against alternative distance measures, specifically cosine similarity and Mahalanobis distance. As shown in Table~\ref{tab:distances}, cosine similarity -- which would be a natural choice for CLIP embeddings -- yields aligned performance with Euclidean distance, which is expected given CLIP embeddings exhibit almost uniform norms, making cosine similarity essentially a monotonic function of Euclidean distance. Conversely, Mahalanobis distance performs worse since covariance estimation becomes numerically unstable for domains with limited samples (fewer than 500 samples), necessitating the exclusion of some adapters and thus degrading performance. Overall, Euclidean distance emerges as the simplest, most robust choice for our method.

\subsection{Full Fine-Tuning (FFT)}
\label{sec:fft_evaluation}
We explore whether our library could be constructed using fully fine-tuned models instead of LoRA adapters. 
Table \ref{tab:fft_vs_lora} reports the results achieved either by deploying and fusing fully fine-tuned models or LoRA adapters in our library. While aggregating fully fine-tuned models is a known practice to merge different knowledge -- as explored in ~\cite{li2024training} -- the results indicate no benefits over our LoRA-based approach. Moreover, storing and merging full models is significantly more computationally expensive than operating with adapters, introducing a sizable overhead at inference time. Full fine-tuning is more prone to overfitting, especially on smaller datasets, whereas LoRA adapters are lightweight and can be trained effectively with limited data. This reinforces our choice of using LoRA adapters, which are modular, efficient, and easily combinable.

\subsection{Domain Navigators: DINO vs. CLIP}
\label{sec:dino_vs_clip}

\methodname uses CLIP~\cite{radford2021learning} to navigate into the LoRA Library and pick the most relevant adapters to combine. However, different visual encoders could serve the same purpose. 
In Table~\ref{tab:dino_vs_clip}, we test the use of an alternative domain navigator -- DINO v2 \cite{oquabdinov2} -- and compare the performance achieved by \methodname variants using this latter or CLIP.

On average, the two perform comparably, with CLIP embeddings slightly outperforming DINO ones in guiding adapter selection on average, likely due to their joint text-image embedding space capturing semantic information more effectively.
Nonetheless, this experiment proves that \methodname is not bound to use CLIP as the domain navigator, although this latter provides nice properties in terms of explainability -- as showcased in Section \ref{sec:blip_lora_labeling}.

\section{Additional Qualitative Results}
\label{sec:additional_qualitative_results}

\begin{figure*}[t]
    \centering
    \begin{subfigure}{0.49\textwidth}
        \centering
        \includegraphics[width=\textwidth]{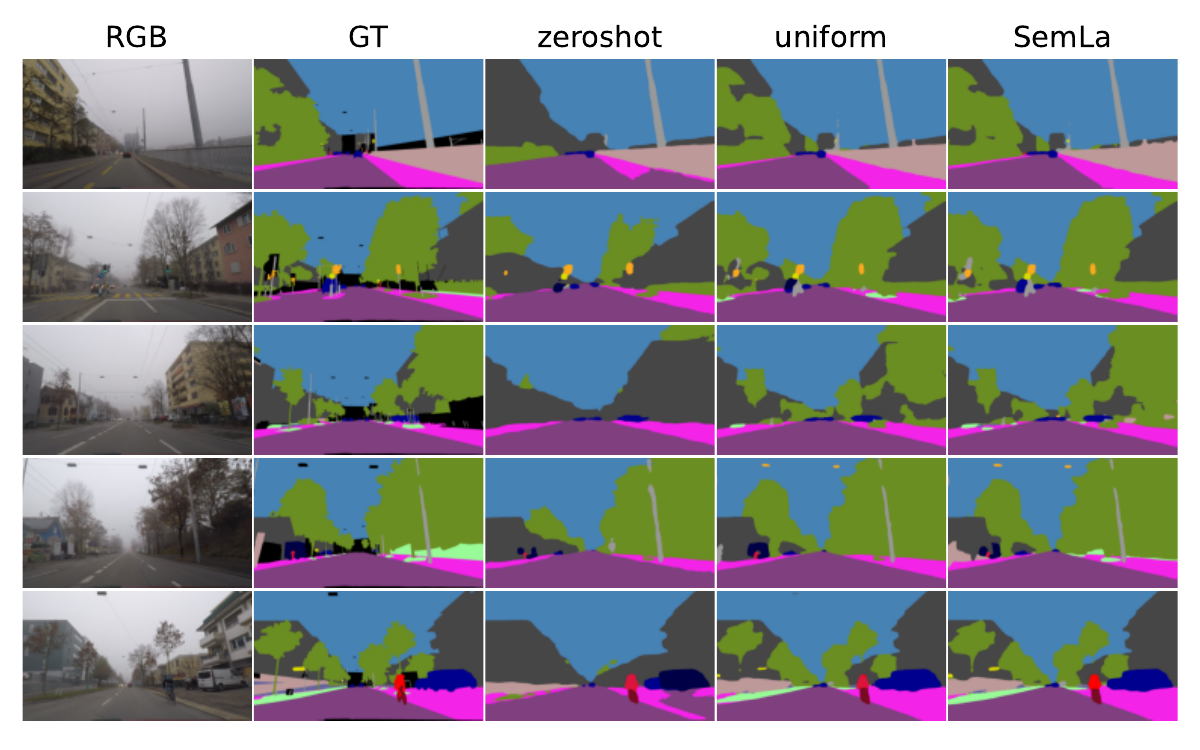}
        \caption{ACDC Fog}
        \label{subfig:acdc_fog}
    \end{subfigure}
    \hfill
    \begin{subfigure}{0.49\textwidth}
        \centering
        \includegraphics[width=\textwidth]{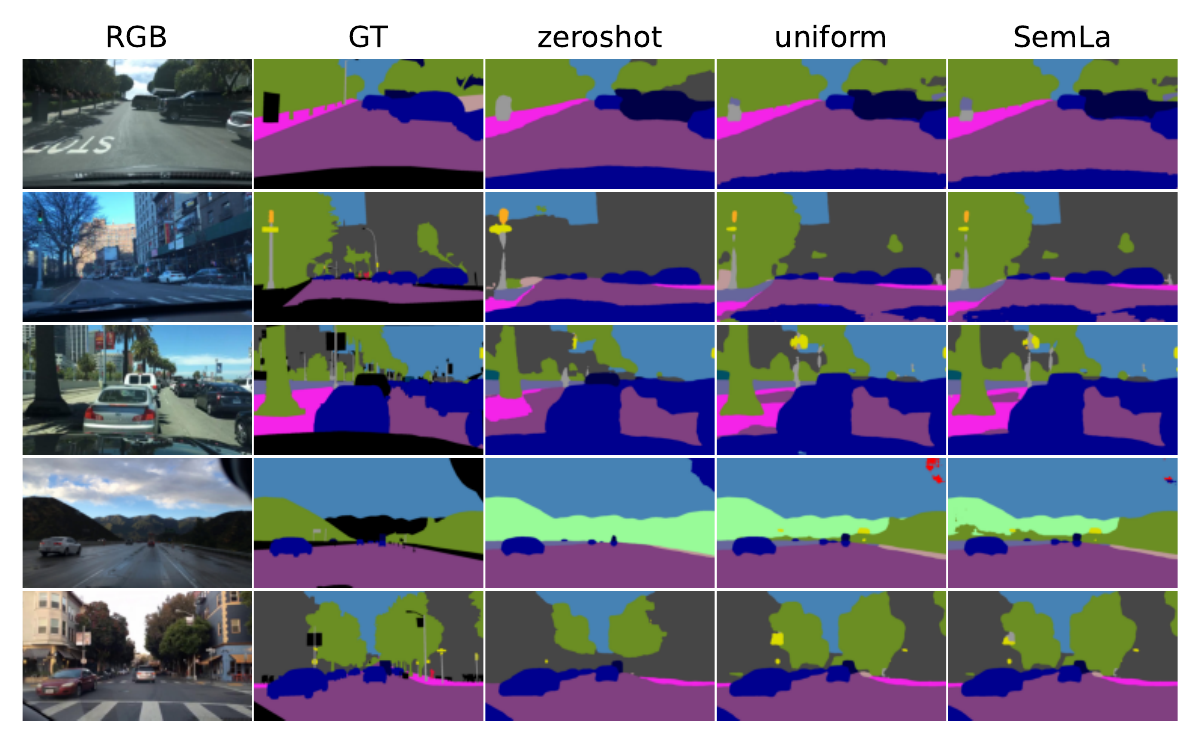}
        \caption{BDD}
        \label{subfig:bdd}
    \end{subfigure}

    \vspace{1em} %

    \begin{subfigure}{0.49\textwidth}
        \centering
        \includegraphics[width=\textwidth]{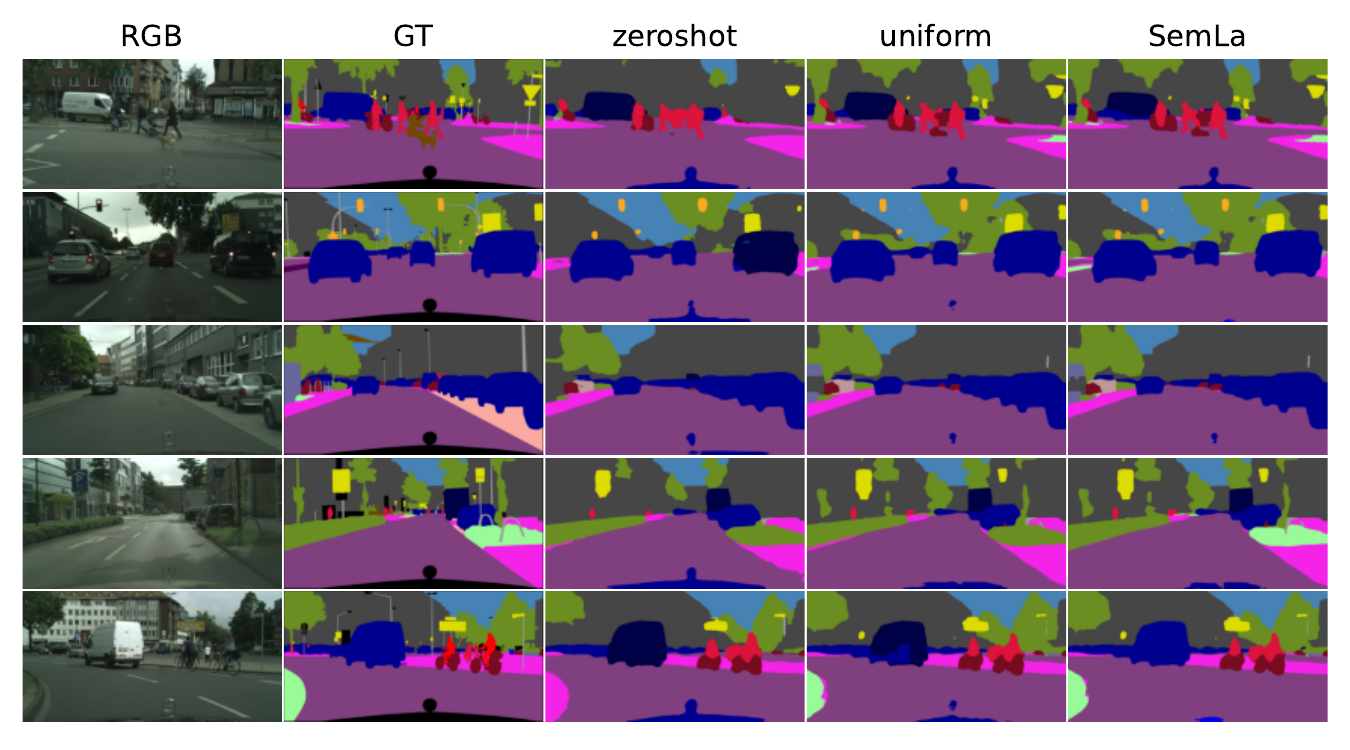}
        \caption{Cityscapes (CS)}
        \label{subfig:cs}
    \end{subfigure}
    \hfill
    \begin{subfigure}{0.49\textwidth}
        \centering
        \vspace{0.7em} %
        \includegraphics[width=0.7\textwidth]{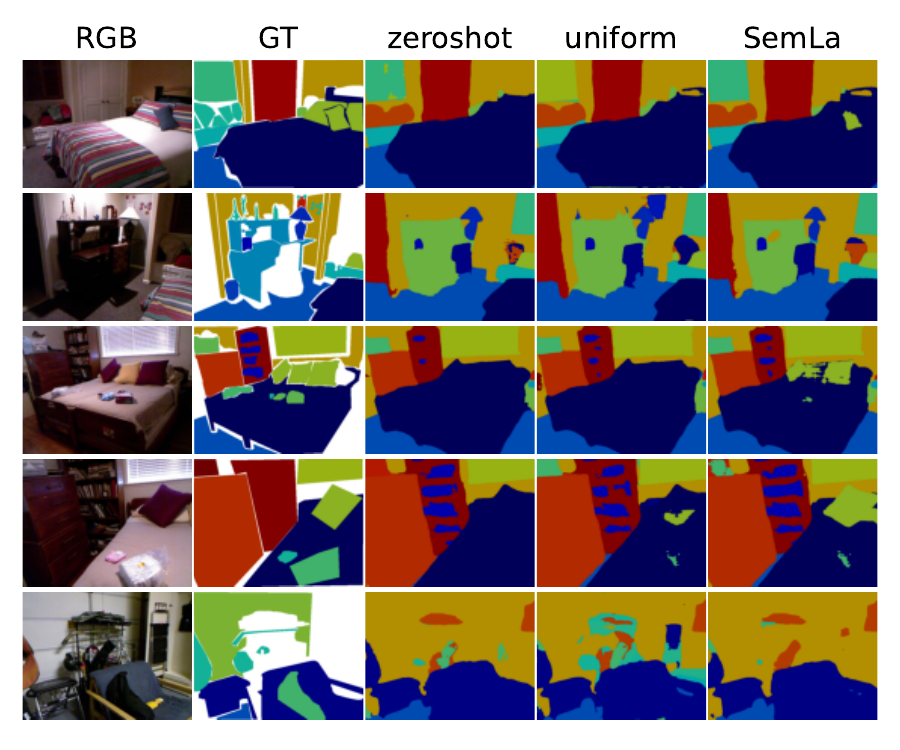}
        \caption{NYU}
        \label{subfig:nyu}
    \end{subfigure}
    \vspace{-0.3cm}
    \caption{\textbf{Additional qualitative results.} The datasets displayed are ACDC Fog, BDD, Cityscapes (CS), and NYU. For each dataset, images are shown in order: Input Image, Zero-Shot, Uniform Merging, \methodname~(Ours), Ground Truth. Our method produces more accurate and detailed segmentations across various domains.}
    \label{fig:qualitative_results_2x2}
\end{figure*}

Figure~\ref{fig:qualitative_results_2x2} presents additional qualitative segmentation results comparing the zero-shot baseline, uniform merging, and \methodname across different domains.
These examples further confirm the effectiveness of our method in adapting to diverse and challenging domains without any additional training being conducted.

\section{Discussion: Real-World Deployment}
\label{sec:discussion}

While \methodname demonstrates strong performance in controlled experimental settings, deploying it in real-world applications introduces additional challenges and considerations. In this section, we discuss practical aspects related to the use of CLIP as a domain navigator and propose strategies to address potential limitations.

\paragraph{CLIP as a Domain Navigator for Specific Domains.}
Although CLIP has shown remarkable generalization capabilities across diverse domains -- as evidenced by our extensive 20-domain benchmark -- it may struggle in niche or highly specialized domains~\cite{wortsman2022robust}. When bringing \methodname into production for such specific use cases, it is important to account for this potential limitation. If the target domain is well-scoped, using a fine-tuned domain navigator, with better semantic understanding, might provide better performance.

Alternatively, a hierarchical approach could be explored: a general CLIP model can provide a coarse understanding of the domain, identifying then a domain-specific CLIP expert. The expert is then tasked with computing the LoRA distances more precisely.

\paragraph{Efficiency in Production Environments.}
In production-intensive applications, dynamically loading and unloading dedicated LoRA adapters for each individual input image may be impractical due to computational overhead. While this overhead is significantly lower than the one introduced by retraining the model at test time -- as required by most traditional test-time adaptation methods -- it is still non-negligible.
For applications that do not require real-time processing, such as batch processing of large volumes of images (e.g., processing data accumulated over 24 hours), a practical approach involves pre-computing the CLIP embeddings for all images. The images can then be clustered based on their embeddings, and a batch centroid can guide the fusion of relevant LoRA adapters for the entire batch. This reduces the frequency of adapter loading and improves efficiency by applying the same fused model to similar images.
In contrast, real-time applications in the field of robotics and autonomous driving cannot rely on batch processing due to their immediate response requirements. In these cases, we propose implementing a debouncing mechanism that triggers adapter swapping only when there is a significant change in the domain. Specifically, the system can monitor the CLIP embeddings of incoming images—or use an exponential moving average (EMA) of these embeddings—and compare them to the embeddings associated with the currently active adapters. If the embedding distance exceeds a predetermined threshold, indicating that a new domain has been encountered, the system triggers the retrieval and fusion of new adapters. This approach ensures that the model adapts only when necessary, minimizing computational overhead while maintaining adaptability.
This strategy is analogous to concepts proposed in domain-adaptive systems like HAMLET~\cite{colomer2023toadapt}, where adaptation occurs only upon detecting domain shifts. Furthermore, in a real-world deployment, the prediction process can be presented with an average LoRA distance metric. As shown in our analysis, this metric provides an additional source of confidence estimation by indicating how well the selected adapters align with the target domain. Such a heuristic contributes to the study of model calibration and can be valuable for downstream tasks—effectively informing whether to trust the model's predictions in critical applications.

\paragraph{Scalability and Model Calibration.}
Scaling \methodname to handle a vast number of adapters introduces challenges in identifying and addressing library weaknesses. Automated strategies for recognizing gaps in the library -- such as monitoring frequent occurrences of high embedding distances -- can prompt the training of new adapters to fill these gaps. Integrating a LoRA support score into the system allows for continuous monitoring of the model's performance relative to the domain coverage of the adapter library. This not only enhances scalability but also improves the system's robustness and reliability in dynamic environments.

%% file: tables/table_blip.tex
\begin{table*}[t]
    \centering
    \setlength{\tabcolsep}{4pt}
    \scalebox{0.49}{
\begin{tabular}{llll}
\toprule
Dataset & Caption: & Question: Where is this? Answer: & Question: Describe The environment in two words? Answer: \\
\midrule
bdd & the view from the driver's seat of a car on a street in san francisco, ca, june 2018 & the city of los angeles, california & The environment in two words is the environment in which it is located. \\
idd & road in kolkata, india, photo by person & a street in bangalore, India 
& city, road, traffic jam \\
nyu & a view of the kitchen in the house i'm renting in san francisco, ca, in summer 2008 & the house i grew up in, in san francisco, california, usa & blue and white \\
acdc-rain & the rain is coming down hard, but the streets are dry, and the cars are moving along the road & berlin, germany, street view, rain & rain \\
acdc-fog & a view from the driver's seat of a car on a highway on a foggy morning in kiev, ukraine & the highway in bordeaux, france, on a foggy day in october 2018 & foggy, rainy, cloudy, misty \\
muses-snow-day & the view from the driver's seat of a car on a city street with buildings in sight & the city of berlin, germany, on a rainy winter day & Rainy day in vienna, austria, with trees and buildings \\
coconutL & person & a small town in the middle of nowhere, nyc, usa & The environment is where the person lives, works, and plays. \\
acdc-night & street at night in kiev, ukraine, with traffic lights and a car on the road & austria & dark and light, city, traffic \\
a150 & the blue house & a house in the middle of the woods & The environment is where the person lives, works, or plays. \\
Cityscapes & street view of berlin, germany & berlin, germany, in the year 2014 & city, street, road, traffic light \\
pc59 & person & the house of the person in the picture & blue sky, green grass \\
muses-fog-day & a view from the driver's seat of a car on a rainy day in bordeaux, france & a foggy day in bordeaux, france, driving on the autoroute & foggy, rainy, misty \\
muses-clear-day & the car driving on the street in the city & berlin, germany, in the year 2040, a virtual reality simulation & city, road, street \\
acdc-snow & street in krasnodar, russia, april 2019 & austria & snow, winter \\
muses-fog-night & the road at night, with car lights visible & the road in the dark, in the middle of nowhere, at night & dark and light \\
muses-clear-night & the car on the road at night, with city lights in the background & austria & night, city, traffic, street lights \\
mv & street view of kuala lumpur, malaysia, with the city's main road visible & australia & urban, city, cityscape \\
muses-snow-night & a view of the city from a car's windshield at night, with city lights and snow visible & a city in the uk, in wintertime, with a car driving on the road & snow, rain, night, city \\
\bottomrule
\end{tabular}
}\vspace{-0.3cm}
\caption{
\textbf{Text generation results using BLIP-2 \cite{li2023blip}.} For the image embedding, the average embedding across all images from each dataset was computed. Then different prompts were given to the model, as presented at the top of the table.
}\vspace{-0.3cm}
\label{tab:blip2}
\end{table*}

%% file: tables/distance_ablations.tex
\begin{table*}[t]
    \centering
    \setlength{\tabcolsep}{3pt}
    \scalebox{0.58}{
     \hspace*{-0.7cm}
     \begin{tabular}{l|cccc|cccccccc|cccccccc|c}
    \hline
    \multirow{2}{*}{Method}
      & \multicolumn{4}{c|}{ACDC}
      & \multicolumn{8}{c|}{MUSES}
      & \multirow{2}{*}{CS}
      & \multirow{2}{*}{BDD}
      & \multirow{2}{*}{MV}
      & \multirow{2}{*}{A150}
      & \multirow{2}{*}{IDD}
      & \multirow{2}{*}{PC59}
      & \multirow{2}{*}{NYU}
      & \multirow{2}{*}{COCONutL*}
      & \multirow{2}{*}{h-mean} \\
    & rain & snow & fog & night
    & clear (d) & clear (n) & rain (d) & rain (n) & fog (d) & fog (n) & snow (d) & snow (n)
    & & & & & & & & \\
    \hline
   Uniform~\cite{li2024training}
      & 67.40 & 66.35 & 69.71 & 49.98 & 58.28 & 55.78 & 54.70 & 45.09 & {73.75} & 45.16 & 61.02 & 49.08 & 62.18 & {58.19} & {31.51} & 37.25 & 38.83 & 63.06 & 48.93 & ({67.62}) & 51.89  \\
      \methodname{} with \textbf{Euclidean} (ours)
      & 67.71 & 68.95 & 71.92 & 51.73 & 61.09 & 60.06 & 57.60 & 47.35 & 72.97 & 52.38 & 67.28 & 55.92 & 63.91 & {57.30} & 31.12 & 38.18 & 40.16 & 64.75 & 51.35 & (67.26) & 54.16  \\
    \methodname with \textbf{Cosine} 
        & 67.76 & 68.67 & 72.52 & 51.24 & 61.55 & 60.22 & 57.76 & 47.03 & 73.03 & 48.10 & 66.82 & 56.67 & 63.53 & 57.52 & 30.24 & 38.10 & 39.80 & 64.65 & 50.93 & (67.31) & 53.70 \\
      \methodname with \textbf{Mahalanobis} $\dagger$ &  59.94 & 63.18 & 67.70 & 45.90 & 56.26 & 50.54 & 48.96 & 36.71 & 75.74 & 34.93 & 56.84 & 35.89 & 57.30 & 56.54 & 30.03 & 37.85 & 39.78 & 64.43 & 50.55 & (67.69) & 47.87  \\
        \hline
    \end{tabular}}\vspace{-0.2cm}
   \caption{\textbf{Ablation study -- Distance Metrics Comparison (CAT-Seg \cite{Cho_2024_CVPR})}. 
 Comparing SemLA with alternative distances. On MUSES, (d) and (n) stand for day and night. ( ) means excluded from h-mean. $\dagger$ For Mahalanobis, source domains where the covariance cannot be computed are excluded from the library. The parameters for Cosine, Mahalanobis, and Late Fusions ( $\tau$ and $K$ ) are tuned independently to achieve the best results with each variant.}
    \label{tab:distances}
\end{table*}

%% file: tables/table_fft.tex
\begin{table*}[t]
    \centering
    \setlength{\tabcolsep}{3pt}
    \scalebox{0.62}{
    \begin{tabular}{l|cccc|cccccccc|cccccccc|c}
    \hline
    \multirow{2}{*}{Method}
      & \multicolumn{4}{c|}{ACDC}
      & \multicolumn{8}{c|}{MUSES}
      & \multirow{2}{*}{CS}
      & \multirow{2}{*}{BDD}
      & \multirow{2}{*}{MV}
      & \multirow{2}{*}{A150}
      & \multirow{2}{*}{IDD}
      & \multirow{2}{*}{PC59}
      & \multirow{2}{*}{NYU}
      & \multirow{2}{*}{COCONutL*}
      & \multirow{2}{*}{h-mean} \\
    & rain & snow & fog & night
    & clear (d) & clear (n) & rain (d) & rain (n) & fog (d) & fog (n) & snow (d) & snow (n)
    & & & & & & & & \\
    \hline
    Zero-shot~\cite{Cho_2024_CVPR}
      & 46.53 & 48.04 & 47.09 & 37.93
      & 44.43 & 39.29 & 38.95 &27.78 & 53.73 & 25.35 & 43.56 & 33.29 & 47.11 & 47.95
      & 25.69 & 37.83 & 35.39 & 63.33 & 49.38 & (68.26)
      & 39.39 \\
    Oracles (LoRA)
      & 70.94 & 69.22 & 69.98 & 51.55 & 69.36 & 57.09 & 54.28 & 52.11 & 75.85 & 61.26 & 66.25 & 54.35 & 67.47 & 60.06 & 49.57 & 53.99 & 64.34 & 68.68 & 61.90 & (70.44) & 61.05 \\
      Oracles (FFT)
      & 70.97 & 72.02 & 73.78 & 53.09 & 70.49 & 58.20 & 55.57 & 53.18 & 74.90 & 62.64 & 65.83 & 58.67 & 70.35 & 61.03 & 50.56 & 52.84 & 66.38 & 68.43 & 64.36 & (68.36) & 62.38 \\
      \hline
    Uniform~\cite{li2024training} (FFT)
      & 69.01 & 67.91 & 73.28 & 51.71 & 61.44 & 57.28 & 54.99 & 43.13 & 74.14 & 36.91 & 57.81 & 52.99 & 62.29 & 58.05 & 30.62 & 36.34 & 39.73 & 62.60 & 48.09 & (65.29) & 51.43 \\
    \methodname{} (FFT)
      & 69.54 & 72.07 & 73.20 & 52.87 & 62.78 & 59.49 & 57.44 & 45.59 & 74.24 & 53.22 & 64.54 & 56.75 & 65.52 & 58.28 & 28.44 & 31.26 & 41.33 & 62.01 & 46.02 & (63.64) & 52.79 \\
       \methodname{} (LoRA)
      & 67.71 & 68.95 & 71.92 & 51.73 & 61.09 & 60.06 & 57.60 & 47.35 & 72.97 & 52.38 & 67.28 & 55.92 & 63.91 & {57.30} & 31.12 & 38.18 & 40.16 & 64.75 & 51.35 & (67.26) & 54.16 \\
    \hline
    \end{tabular}}
    \vspace{-0.2cm}
    \caption{\textbf{Ablation study -- Full Fine-Tuning vs LoRA Adaptation  (CAT-Seg \cite{Cho_2024_CVPR}).} We use full fine-tuned models instead of LoRA adapters and measure the impact on performance over our 20-domain benchmark in leave-one-out setting. On MUSES, (d) and (n) stand for day and night. ( ) means excluded from h-mean.  \methodname{} (LoRA) with $\tau=0.05$, and top-$K=5$; \methodname{} (FFT) with $\tau=0.01$, and top-$K=9$.}
    \label{tab:fft_vs_lora}
\end{table*}

%% file: tables/table_dino.tex
\begin{table*}[t]
    \centering
    \setlength{\tabcolsep}{3pt}
    \scalebox{0.62}{
     \hspace*{-0.6cm}
     \begin{tabular}{l|cccc|cccccccc|cccccccc|c}
    \hline
    \multirow{2}{*}{Method}
      & \multicolumn{4}{c|}{ACDC}
      & \multicolumn{8}{c|}{MUSES}
      & \multirow{2}{*}{CS}
      & \multirow{2}{*}{BDD}
      & \multirow{2}{*}{MV}
      & \multirow{2}{*}{A150}
      & \multirow{2}{*}{IDD}
      & \multirow{2}{*}{PC59}
      & \multirow{2}{*}{NYU}
      & \multirow{2}{*}{COCONutL*}
      & \multirow{2}{*}{h-mean} \\
    & rain & snow & fog & night
    & clear (d) & clear (n) & rain (d) & rain (n) & fog (d) & fog (n) & snow (d) & snow (n)
    & & & & & & & & \\
    \hline
    Zero-shot~\cite{Cho_2024_CVPR}
      & 46.53 & 48.04 & 47.09 & 37.93
      & 44.43 & 39.29 & 38.95 &27.78 & 53.73 & 25.35 & 43.56 & 33.29 & 47.11 & 47.95
      & 25.69 & 37.83 & 35.39 & 63.33 & 49.38 & (68.26)
      & 39.39 \\
    Oracles 
      & 70.94 & 69.22 & 69.98 & 51.55 & 69.36 & 57.09 & 54.28 & 52.11 & 75.85 & 61.26 & 66.25 & 54.35 & 67.47 & 60.06 & 49.57 & 53.99 & 64.34 & 68.68 & 61.90 & (70.44) & 61.05  \\
      \hline
     Uniform~\cite{li2024training}
      & 67.40 & 66.35 & 69.71 & 49.98 & 58.28 & 55.78 & 54.70 & 45.09 & \ {73.75} & 45.16 & 61.02 & 49.08 & 62.18 & \ {58.19} & \ {31.51} & 37.25 & 38.83 & 63.06 & 48.93 & ({67.62}) & 51.89 \\
    \methodname{} (DINOv2)
      & 68.40 & 68.26 & 73.57 & 51.18 & 61.94 & 59.58 & 56.06 & 48.43 & 73.81 & 52.60 & 67.42 & 56.33 & 64.04 & 58.23 & 31.02 & 37.45 & 40.12 & 64.40 & 50.52 & (67.63) & 54.14 \\
       \methodname{} (CLIP)
      & 67.71 & 68.95 & 71.92 & 51.73 & 61.09 & 60.06 & 57.60 & 47.35 & 72.97 & 52.38 & 67.28 & 55.92 & 63.91 & {57.30} & 31.12 & 38.18 & 40.16 & 64.75 & 51.35 & (67.26) & 54.16 \\
    \hline
    \end{tabular}}\vspace{-0.2cm}
   \caption{\textbf{Ablation study -- CLIP~\cite{radford2021learning} vs DINOv2~\cite{oquabdinov2} for domain navigation (CAT-Seg \cite{Cho_2024_CVPR}).} -- We generate the weights for merging the LoRAs based on features extracted from DINOv2 or CLIP, and evaluate the impact on performance over our 20-domain benchmark in a leave-one-out setting. On MUSES, (d) and (n) stand for day and night. ( ) means excluded from h-mean.}
    \label{tab:dino_vs_clip}
\end{table*}